\documentclass[final,1p,times,twocolumn]{elsarticle}

\usepackage{lineno}

\usepackage{multirow}
\usepackage{threeparttable}
\usepackage{amsmath}
\usepackage{amssymb}

\usepackage{graphicx}
\usepackage{comment}
\usepackage{cases}
\usepackage{color}
\usepackage{subfigure}
\usepackage{csquotes}
\newcommand*\rot{\rotatebox[origin=c]{90}}
\usepackage{multirow}
\usepackage{booktabs}

\newcommand*{\twoelementtable}[3][l]%
{%
\renewcommand{\arraystretch}{0.8}%
\begin{tabular}[t]{@{}#1@{}}%
        #2\tabularnewline
        #3%
\end{tabular}%
}
\usepackage{array}
\usepackage{tabularx}

\usepackage{mathtools}

\journal{Pattern Recognition}

\begin{document}

\begin{frontmatter}

\title{MLAN: Multi-Level Adversarial Network for Domain Adaptive Semantic Segmentation} 

\author{Jiaxing Huang}
\ead{jiaxing.huang@ntu.edu.sg}

\author{Dayan Guan}
\ead{dayan.guan@ntu.edu.sg}

\author{Aoran Xiao}
\ead{aoran.xiao@ntu.edu.sg}

\author{Shijian Lu\corref{cor1}}
\ead{shijian.lu@ntu.edu.sg}
\cortext[cor1]{Corresponding author}

\address{Nanyang Technological University, 50 Nanyang Avenue, Singapore 639798}

\begin{abstract}
Recent progresses in domain adaptive semantic segmentation demonstrate the effectiveness of adversarial learning (AL) in unsupervised domain adaptation. However, most adversarial learning based methods align source and target distributions at a global image level but neglect the inconsistency around local image regions. This paper presents a novel multi-level adversarial network (MLAN) that aims to address inter-domain inconsistency at both global image level and local region level optimally. MLAN has two novel designs, namely, region-level adversarial learning (RL-AL) and co-regularized adversarial learning (CR-AL). Specifically, RL-AL models prototypical regional context-relations explicitly in the feature space of a labelled source domain and transfers them to an unlabelled target domain via adversarial learning. CR-AL fuses region-level AL and image-level AL optimally via mutual regularization. In addition, we design a multi-level consistency map that can guide domain adaptation in both input space ($i.e.$, image-to-image translation) and output space ($i.e.$, self-training) effectively. Extensive experiments show that MLAN outperforms the state-of-the-art with a large margin consistently across multiple datasets.
\end{abstract}


\end{frontmatter}

\section{Introduction}

Semantic segmentation enables pixel-wise scene understanding, which is crucial to many real-world applications such as autonomous driving. The recent surge of deep learning methods has significantly accelerated the progress in semantic segmentation \cite{chen2017deeplab,long2015fully} but at the price of large-scale per-pixel labelled datasets \cite{cordts2016cityscapes} which are often prohibitively costly to collect in terms of time and money. Latest progresses in computer graphics engines provide a possible way to circumvent this problem by synthesizing photo-realistic images together with pixel-wise labels automatically \cite{richter2016playing,ros2016synthia}. However, deep neural networks trained with such synthetic image data tend to perform poorly on real image data due to clear domain gaps \cite{vu2019advent}.

\begin{figure}[!t]
\centering 
\includegraphics[width=.98\linewidth]{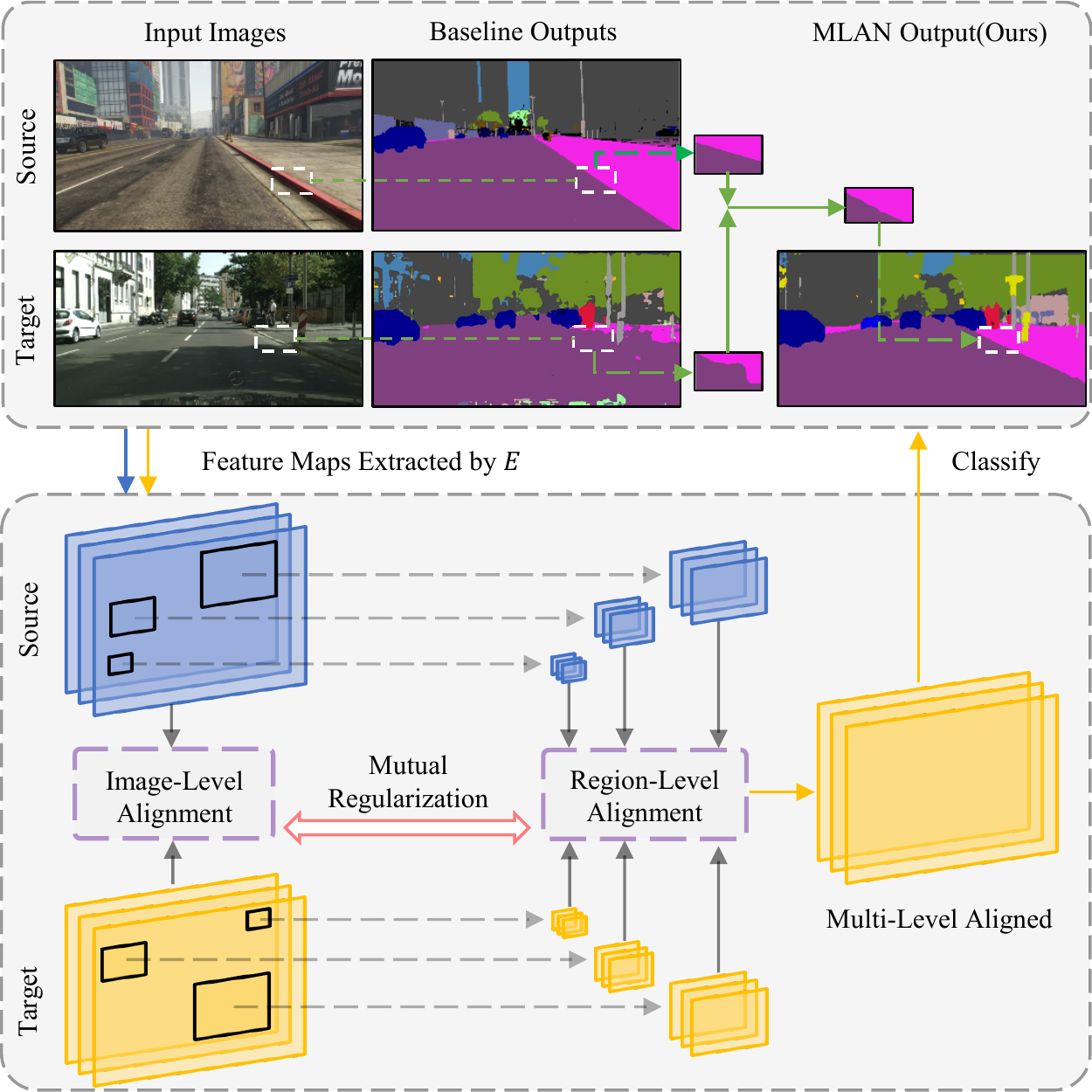}
\caption{Our proposed multi-level adversarial network (MLAN) improves domain adaptive semantic segmentation: MLAN adapts features at multiple levels for joint global image-level and local region-level alignment between source and target domains as illustrated in the bottom part. Arrows and boxes in blue and yellow denote source and target data-flows, while dash and solid arrows in gray represent alignment processes. The segmentation outputs with and without our proposed technique are illustrated in the top part (white boxes and green arrows highlight local image regions). Best viewed in color.
}
\label{fig:intro}
\end{figure}

Unsupervised domain adaptation (UDA) has been introduced to address the domain bias/shift issue, and most existing unsupervised domain adaptation techniques \cite{tsai2018learning,vu2019advent,luo2019taking,wang2020differential} employ adversarial learning to align the data representations of source and target domains via a discriminator \cite{tzeng2017adversarial,tsai2018learning,luo2019taking,tsai2019domain,vu2019advent}. In these approaches, the adversarial loss is essentially a binary cross-entropy which is evaluated according to whether the generated representation is from the source or target domain. However, the global image-level domain classification signal does not have sufficient capacity to capture local/regional features and characteristics ($e.g.$, consistency of region-level context-relations) as illustrated in Figure \ref{fig:intro}, which is usually too broad and weak to handle the pixel-level semantic segmentation task. 

In \cite{huang2020contextual}, we investigated domain adaptive semantic segmentation that addresses the limitation of global image-level alignment by exploiting local region-level consistency across domains. Specifically, \cite{huang2020contextual} first learns prototypical region-level context-relations explicitly in source domains (with synthetic images) and then transfers the learnt region-level context-relations to target domains (with real images) via adversarial learning. 
Beyond region-level adversarial learning in \cite{huang2020contextual} (to be described in Section 3.2), this work further investigates how to exploit consistency and co-regularization of local and global information for optimal domain adaptive segmentation. Specifically, we design a multi-level adversarial networks (MLAN) and co-regularized adversarial learning (CR-AL) that achieve both global image-level and local region-level alignment optimally as illustrated in Figure \ref{fig:intro} (to be described in Section 3.3). MLAN employs a mutual regularizer that uses local and global consistencies to coordinate global image-level adversarial learning (IL-AL) and local region-level adversarial learning (RL-AL). In particular, local consistencies regularize IL-AL by enforcing region-level constraints during image-level alignment, e.g., it increases IL-AL loss for regions with large region-level inconsistency. Similarly, global consistencies regularize RL-AL by enforcing image-level constraints during region-level alignment, e.g., it increases RL-AL loss for hard images with large domain gaps. The multi-level and mutually regularized adversarial learning adds little overhead during inference but just four one-layer classifiers during the training stage.

In addition, we introduce a multi-level consistency map (MLCM) that guides domain adaptive segmentation in image translation in the input space and self-training in the output space (to be described in Section 3.4). For input-space adaptation, MLCM learns adaptive image-to-image translation that can attend to specific image regions with large domain gaps. For output-space adaptation, MLCM introduces domain gap information and encourages self-training to select samples/pixels that have larger domain gaps in pseudo labeling. This helps prevent over-fitting to easy samples/pixels with smaller domain gaps effectively. Extensive experiments demonstrate the effectiveness of proposed two new designs, and MLAN surpasses the state-of-the-art including our prior work \cite{huang2020contextual} consistently by large margins.

The organization of this paper is as follows:  We  start by reviewing related work on unsupervised domain adaptive semantic segmentation in Section 2. Then, Sections 3 details our proposed MLAN for domain adaptive semantic segmentation. We provide both qualitative and quantitative evaluations in Section 4, and conclude this paper in Section 5.

\section{Related work}
Unsupervised Domain Adaptation (UDA) ~\cite{deng2018active,liang2019exploring} aims to learn a model from source supervision only that can perform well in target domains. It has been extensively explored for the tasks of classification~\cite{li2018adaptive,zuo2020challenging,rahman2020correlation} and detection~\cite{chen2018domain,xu2020exploring}. In recent years, UDA for semantic segmentation \cite{hoffman2016fcns} has drawn increasing attention as training high-quality segmentation model requires large-scale per-pixel annotated dataset, which is often prohibitively expensive and labor-intensive. 
Most existing unsupervised domain adaptive semantic segmentation methods can be broadly classified into three categories including adversarial learning based \cite{tsai2018learning,vu2019advent,luo2019taking,guan2020scale}, image translation based \cite{hoffman2018cycada,yang2020fda}, and self-training based \cite{zou2018unsupervised,zou2019confidence}.

\textbf{Adversarial learning based segmentation} minimizes domain divergences by matching the marginal distributions between source and target domains. Hoffman \textit{et al.} \cite{hoffman2016fcns} first proposed a domain adaptive segmentation method by employing a discriminator to globally align source and target features and introducing a constraint from category-level statistic to guarantee the semantic similarity. Chen et al. \cite{chen2017no} implemented class-wise adversarial learning and grid-wise soft label transferring to achieve global and class-wise alignment. Tsai \textit{et al.} \cite{tsai2018learning} demonstrated the spatial similarities across domains and applied adversarial learning in output space. Vu \textit{et al.} \cite{vu2019advent} proposed an adversarial entropy minimization method to achieve structure adaptation in entropy space. Lou \textit{et al.} \cite{luo2019taking} designed a category-level adversarial learning technique to enforce semantic consistency in marginal distribution alignment from source domain to target domain. Guan \textit{et al.} \cite{guan2020scale} proposed a scale-invariant constraint to regularize adversarial learning loss for preserving semantic information. 
 
\textbf{Image translation based segmentation} reduces domain gap in the input space by transferring the style of source-domain images to target domain \cite{hoffman2018cycada,huang2021fsdr,li2019bidirectional,yang2020fda}. Hoffman \textit{et al.} \cite{hoffman2018cycada} employs generative adversarial networks to produce target-like source images for direct input adaptation. Li \textit{et al.} \cite{li2019bidirectional} alternatively trained image translation networks and segmentation model in a bidirectional learning manner. Yang \textit{et al.} \cite{yang2020fda} proposed a simple but effective image translation method for domain adaption by swapping the low-frequency spectrum of source images with target images. 

\textbf{Self-training based segmentation} retrains source-domain models iteratively by including target-domain samples that have high-confident predictions of their pseudo labels (by newly trained source-domain models) \cite{zou2018unsupervised,huang2021cross,zou2019confidence,wang2020differential}. Zou \textit{et al.} \cite{zou2018unsupervised} proposed a class-balanced self-training approach to optimize pseudo-label generation via decreasing the influence of dominant category.
Zou \textit{et al.} \cite{zou2019confidence} refined psudo-label generation and model training by a label entropy regularizer and an output smoothing regularizer respectively.
Wang \textit{et al.}  \cite{wang2020differential} optimized self-training by minimizing the distance of background and foreground features across domain.

This paper presents a multi-level adversarial network (MLAN) that introduces local region-level adversarial learning (RL-AL) and co-regularized adversarial learning (CR-AL) for optimal domain adaptation. In addition, MLAN computes a multi-level consistency map (MLCM) to guide the domain adaptation in both input and output spaces. To our knowledge, this is the first method that addresses domain adaptive semantic segmentation by mutual regularization of adversarial learning at multiple levels effectively.

\section{Proposed methods}
In this section, we present the proposed multi-level adversarial network (MLAN) that aims to achieve both local and global consistencies optimally. MLAN consists of three key components including local region-level adversarial learning (RL-AL), co-regularized adversarial learning (CR-AL), and consistency-guided adaptation in the input and output spaces, which will be described in details in the ensuing subsections 3.2, 3.3, and 3.4, respectively. 

\subsection{Problem definition}
We focus on the problem of unsupervised domain adaptation (UDA) in semantic segmentation. Given the source data $X_{s} \subset \mathbb{R}^{H \times W \times 3}$ with C-class pixel-level segmentation labels $Y_{s} \subset (1,C)^{H \times W}$ ($e.g.$, synthetic scenes from computer graphics engines) and the target data $X_{t} \subset \mathbb{R}^{H \times W \times 3}$ without labels ($i.e.$, real scenes), our objective is to learn a segmentation 
network $F$ that performs on target dataset $X_{t}$ optimally. Existing adversarial learning based domain adaptive segmentation networks heavily rely on the discriminator to align source and target distributions through two learning objectives: supervised segmentation loss on source images and adversarial loss for target-to-source alignment. Specifically, for the source domain, these approaches learn a segmentation model $F$ with supervised segmentation loss. And then, for target domain, these adversarial learning based UDA networks train $F$ to extract domain-invariant features though the mini-max gaming between segmentation model $F$ and a discriminator $D$. Therefore, they formulate this UDA task as:

\begin{equation}
\mathcal{L}(X_{s}, X_{t}) = \mathcal{L}_{seg}(F) + \mathcal{L}_{adv}(F, D)
\end{equation}

However, the current adversarial learning based UDA networks have two crucial limitations. First, they focus on global image-level alignment but neglect local region-level alignment. Second, the global image-level alignment could impair local consistencies even if it is achieved perfectly. The reason is that the discriminator in global image-level alignment takes a whole map as input but outputs a binary domain label only. Consequently, certain local regions that have been well aligned across domains might be deconstructed by the global image-level adversarial loss from the discriminator. We observe such "lack of local region-level consistency" that is essential to semantic segmentation with pixel-level prediction.
\begin{figure}[!t]
\centering
\includegraphics[width=.98\linewidth]{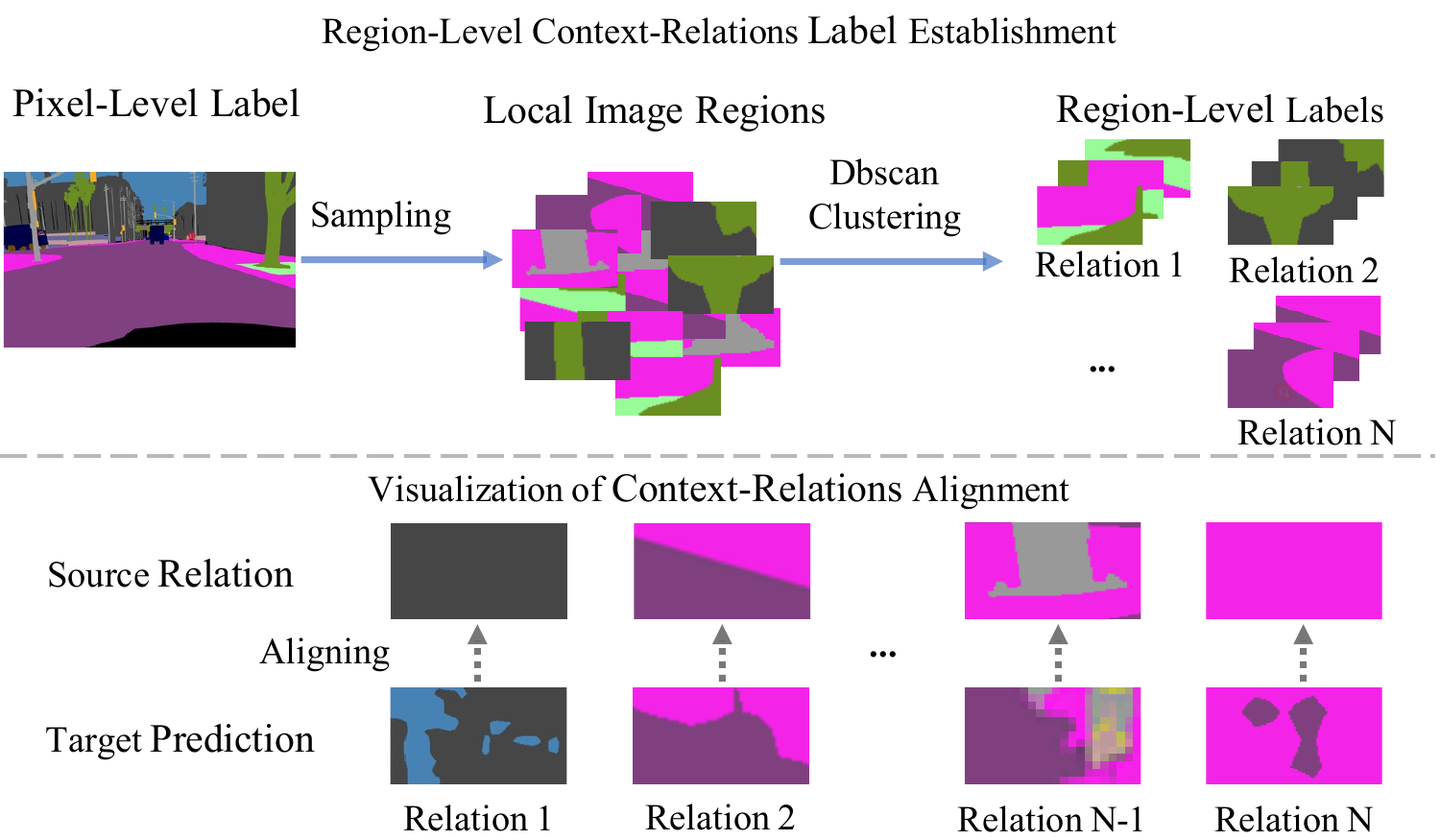}
\caption{Illustration of establishment of region-level context-relation labels: we first sample local image regions from pixel-level segmentation maps (in the provided ground-truth annotations) in the source domain and then implement DBSCAN clustering (based on the histogram of gradient) to assign each local region with an indexed/labelled context-relation. The bottom part visualizes the local region-level alignment that enforces target predictions to have source-like context-relations via adversarial learning.}
\label{fig:local}
\end{figure}

\subsection{Region-level adversarial learning}
This subsection introduces our local region-level adversarial learning (RL-AL) \cite{huang2020contextual} that consists of region-level context-relations discovery and transfer, as shown in Figure \ref{fig:local}.

\textbf{Region-level context-relations label establishment.} In order to implement local region-level adversarial learning (RL-AL), we first explicitly discover and model the region-level context-relations in labelled source domain. Specifically, we first sample patches on the pixel-level ground-truth of source data and then employ Density-based spatial clustering of applications with noise (Dbscan) to cluster them into subgroups based on the histogram of gradient and assign each patch a certain index label ($i.e.$, relation 1, 2, ...., N). We apply this region-level context-relations label establishment process with three patch sizes for explicit local context-relations discovery.  These region-level context-relations labels can 
support our model to conduct alignment at region-level.

\begin{figure}[!t]
\centering
\includegraphics[width=.98\linewidth]{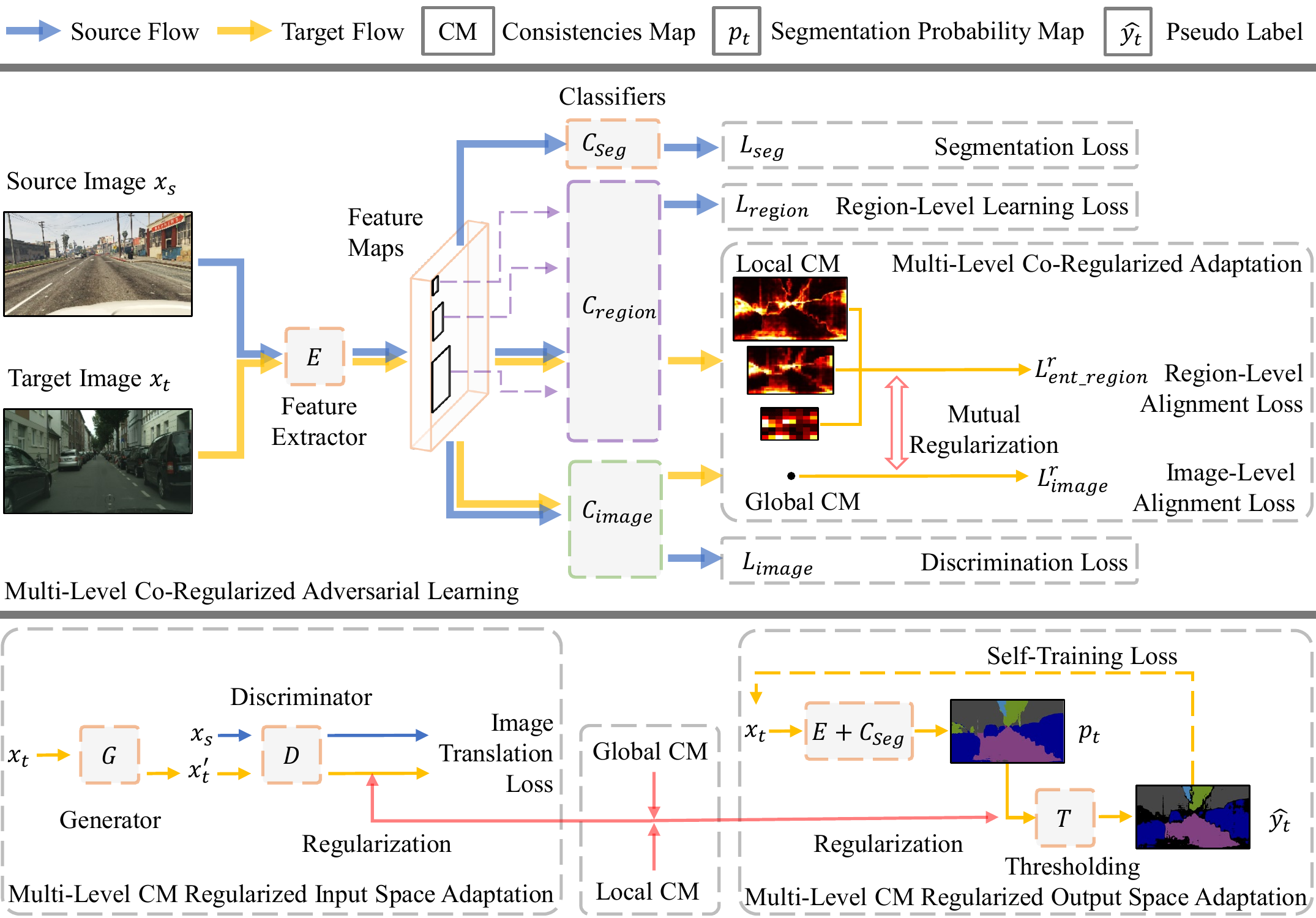}
\caption{Overview of our proposed multi-level adversarial network (MLAN): 
Given input images, feature extractor $E$ extracts features and feeds them to region-level context-relations classifiers $C_{region}$ for classification at three region levels.
In the source flow (highlighted by arrows in blue), $\mathcal{L}_{seg}$ is a supervised segmentation loss on labelled source domain; $\mathcal{L}_{region}$ is a supervised region-level learning loss on source domain with the region-level context-relations labels established in Figure \ref{fig:local}.
In the target flow (highlighted by arrows in yellow), $\mathcal{L}^{r}_{ent\_region}$ is an entropy max-min based unsupervised region-level alignment loss to enforce target predictions to have source-like region-level context-relations. 
$C_{image}$ is the domain classifier that aligns source and target distributions at image-level via loss $\mathcal{L}_{image}$/$\mathcal{L}^{r}_{image}$.
The co-regularized adversarial learning (CR-AL) is achieved by mutually rectifying the region-level and image-level alignments. In addition, the local ($i.e.$, region-level) and global ($i.e.$, image-level) consistencies map (CM) are integrated into a multi-level CM to guide the input space and output space adaptation.
}
\label{fig:framework}
\end{figure}

\textbf{Adaptive entropy max-minimizing for region-level adversarial learning.} As shown in Figure \ref{fig:framework}, $C_{region}$ learns prototypical region-level contextual-relations ($e.g.$, sidewalk-road, building-sky, car-road, etc.) via the supervised learning in labelled source domain ($i.e.$, the established local region-level context-relations label) and transfers the learnt prototypical contextual-relations on unlabelled target data by implementing adaptive entropy max-minimization based adversarial learning.

\textbf{Source flow.} In our region-level adversarial learning (RL-AL), the source data contributes to $L_{seg}$ and $L_{region}$. Given a source image $x_{s} \in X_{s}$, its corresponding segmentation label $y_{s} \in Y_{s}$ and contextual-relation pseudo-label $y_{s\_region} \in Y_{s\_region}$, $p_{s}^{(h, w, c)} = C_{seg}(E(x_{s}))$ is the predicted pixel-level probability map for semantic segmentation; $p_{s\_region}^{(i, j, n)} = C_{region}(E(x_{s}))$ is the predicted region-level probability map for local contextual-relations classification. Thus, the two supervised learning objectives is to
minimize $L_{seg}$ and $L_{region}$, respectively, which are formulated as:

\begin{equation}
\mathcal{L}_{seg}(E, C_{seg}) = \sum_{h, w} \sum_{c} -y_{s}^{(h, w, c)} \log p_{s}^{(h, w, c)}
\end{equation}

\begin{equation}
\mathcal{L}_{region}(E, C_{region}) = \sum_{i, j} \sum_{n} -y_{s\_region}^{(i, j, n)} \log p_{s\_region}^{(i, j, n)}
\end{equation}

\textbf{Target flow.} As the target data is not annotated, we propose an adaptive entropy max-minimization based adversarial training between feature extractor $E$ and classifier $C_{region}$ to enforcing the target segmentation to have source-like context-relations. Given a target image $x_{t} \in X_{t}$, $p_{t\_region}^{(i, j, n)} = C_{region}(E(x_{t}))$ is the predicted region-level probability map for local contextual-relations classification. The entropy based loss $L_{ent\_region}$ is formulated as:

\begin{equation}
\mathcal{L}_{ent\_region}(E, C_{region}) = - \frac{1}{N}\sum_{i, j} \sum_{n} \text{max}\{p_{t\_region}^{(i, j, n)} \log p_{t\_region}^{(i, j, n)} - \mathcal{R}(p_{t\_region}^{(i, j, n)}), 0\},
\label{equ:local}
\end{equation}
where the adaptive optimization is achieve by $\mathcal{R}(p)=\text{average}\{ p\log p\} \times \lambda _{R}$; $\lambda_{R} = (1 - \frac{iter}{max\_iter})^{power}$ with ${power} = 0.9$ is a weight factor decreases with training iteration.

We employ the gradient reverse  layer~\cite{ganin2015unsupervised} for local region-level adversarial learning and the training objective is formulated as:

\begin{equation}
\begin{split}
& \min_{\theta_{E}} \mathcal{L}_{seg} + \lambda_{region}\mathcal{L}_{region} + \lambda_{ent} \mathcal{L}_{ent\_region}, \\
& \min_{\theta_{C_{seg}}} \mathcal{L}_{seg}, \\
& \min_{\theta_{C_{region}}} \lambda_{region}\mathcal{L}_{region} - \lambda_{ent} \mathcal{L}_{ent\_region}, \\
\end{split}
\end{equation}
where $\lambda_{ent}$ is a weight factor to balance the target unsupervised domain adaptation loss and the source supervised loss; $\lambda_{region}$ is a weight factor to balance the local region-level and pixel-level supervised learning on source domain.

\subsection{Co-regularized adversarial learning}
This subsection introduces our co-regularized adversarial network (CR-AL) that integrates local region-level AL with global image-level AL and conducts mutual regularization, as shown in the top part of Figure \ref{fig:framework}.

\textbf{Image-level adversarial learning (IL-AL).} For IL-AL, We employ a domain classifier $C_{image}$ that takes the source and target feature maps as inputs, and predicts domain labels for them ($i.e.$, 0/1 for source/target domain). The global image-level alignment is conducted through the mini-max gaming between this domain classifier $C_{image}$ and feature extractor $E$. The global image-level adversarial learning loss is defined as:

\begin{equation}
\mathcal{L}_{image}(E, C_{image}) = \sum_{u, v} \mathbb{E}[\log C_{image}(E(x_{s}))] + \mathbb{E}[1-\log C_{image}(E(x_{t}))],
\end{equation}
where $C_{image}(E(x_{s})) \in \mathbb{R}^{U \times V}$ is a domain label prediction map with size $3 \times 3$ ($i.e.$, $U=V=3$).

\textbf{Consistencies map (CM) calculation.} With both region-level and image-level ALs, we calculate the region-level consistencies map (RLCM) and image-level consistencies map (ILCM) for mutual regularization between local region-level and global image-level adversarial learning:
\begin{equation}
\begin{split}
&\mathcal{M}_{region} = - \frac{1}{N} \sum_{n} max\{p_{t\_region}^{(i, j, n)} \log p_{t\_region}^{(i, j, n)} - \mathcal{R}(p_{t\_region}^{(i, j, n)}), 0\},\\
&\mathcal{M}_{image} = \mathbb{E}[1 - \log C_{image}(E(x_{t}))],\\
\end{split}
\label{local+global}
\end{equation}
where $p_{t\_region}^{(i, j, n)}$ and $\mathcal{R}(\cdot)$ are described in Equation. \ref{equ:local}.

\textbf{Multi-level mutual regularization.} Then the local consistencies regularized image-level AL and global consistencies regularized region-level AL are formulated as:

\begin{equation}
\begin{split}
\mathcal{L}_{ent\_region}^{r}(E, C_{region}) = & - \frac{1}{N}\sum_{i, j} \sum_{n}\\ &\text{Max}\{p_{t\_region}^{(i, j, n)} \log p_{t\_region}^{(i, j, n)} - \mathcal{R}(p_{t\_region}^{(i, j, n)}), 0\} \times \mathcal{M}_{image},
\end{split}
\label{Rlocal}
\end{equation}
where $p_{t\_region}^{(i, j, n)}$ and $\mathcal{R}(\cdot)$ are described in Equation. \ref{equ:local}; $\mathcal{M}_{image}$ is up-sampled to match the size of $\text{Max}\{p_{t\_region}^{(i, j, n)} \log p_{t\_region}^{(i, j, n)} - \mathcal{R}(p_{t\_region}^{(i, j, n)}), 0\} \in \mathbb{R}^{I \times J}$ during calculation.

\begin{equation}
\begin{split}
\mathcal{L}_{image}^{r}(E, C_{image}) =& \sum_{u, v} \mathbb{E}[\log C_{image}(E(x_{s}))]\\
&+ \sum_{i, j} \mathbb{E}[1-\log C_{image}(E(x_{t}))] \times \mathcal{M}_{region},\\
\end{split}
\label{Rglobal}
\end{equation}
where $\mathbb{E}[1-\log C_{image}(E(x_{t}))]$ is up-sampled to match the size of $\mathcal{M}_{region} \in \mathbb{R}^{I \times J}$ during calculation.

\textbf{Learning objective.} We integrate Equation. \ref{Rlocal} and \ref{Rglobal} to form a co-regularized adversarial learning and the training objective is formulated as:

\begin{equation}
\begin{split}
& \min_{\theta_{E}} \mathcal{L}_{seg} + \lambda_{region}\mathcal{L}_{region} + \lambda_{ent}\mathcal{L}^{r}_{ent\_region} + \lambda_{image} \mathcal{L}^{r}_{image} \\
& \min_{\theta_{C_{seg}}} \mathcal{L}_{seg}\\
& \min_{\theta_{C_{region}}} \lambda_{region}\mathcal{L}_{region} - \lambda_{ent} \mathcal{L}^{r}_{ent\_region}\\
& \max_{\theta_{C_{image}}} \lambda_{image} \mathcal{L}^{r}_{image}\\
\end{split}
\end{equation}
where $\lambda_{region}$, $\lambda_{image}$ and $\lambda_{ent}$ are the weight factors to balance the objectives different losses.

\subsection{Multi-level consistencies regularized input/output adaptation}
This subsection introduces the multi-level consistencies map (MLCM) that guides and rectifies the input space ($e.g.$, image translation) and output space ($e.g.$, self-training) domain adaptation effectively, as shown in the bottom part of Figure \ref{fig:framework}.

The multi-level consistencies map (MLCM) is an incorporation of local region-level consistencies map (RLCM) and global image-level consistencies map (ILCM), which is formulated as:

\begin{equation}
\mathcal{M}_{multi} = \mathcal{M}_{region} \times \mathcal{M}_{image},
\end{equation}
where $\mathcal{M}_{region}$ and $\mathcal{M}_{image}$ are defined in Equation. \ref{local+global}; $\mathcal{M}_{image}$ is up-sampled to match the size of $\mathcal{M}_{region} \in \mathbb{R}^{I \times J}$ during calculation.

We then utilize the calculated MLCM to regularizing the image translation for input space domain adaptation, as shown in the left bottom part of Figure \ref{fig:framework}. MLCM enables adaptive image translation that learns translation with adaptive attention on different regions/images ($i.e.$, focus more on large domain gap regions/images). The training objective of MLCM regularized image translation (MLCMR-IT) is defined as: 

\begin{equation}
\min_{\theta_{G}} \max_{\theta_{D}} \mathcal{L}^{r}_{IT}(G, D) =\mathbb{E}[\log D(X_{s})] + \mathbb{E}[1-\log D(G(X_{t}))] \times \mathcal{M}_{multi},
\end{equation}
where $G$ is the generator; $D$ is the discriminator; the original image translation loss $\mathcal{L}_{IT}(G, D)$ is the same as the regularized loss $\mathcal{L}^{r}_{IT}(G, D)$ without multi-level regularization ($i.e.$, $\mathcal{M}_{multi}$).

Similarly, MLCM is also used to regularize the self-training for output space domain adaptation, as shown in the right bottom part of Figure \ref{fig:framework}. MLCM introduces domain gap information and encourages self-training to select samples/pixels with larger domain gap as pseudo label for preventing over-fitting on easy samples/pixels ($i.e.$, samples with smaller domain gap). The MLCM regularized pseudo label generation is defined as:

\begin{equation}
\hat{y_{t}}^{(h, w)} = \arg\max_{c \in C} \text{\usefont{U}{bbm}{m}{n}1}_{[(p_{t}^{(c)} \times \mathcal{M}_{multi}) > \exp(-k_{c})]}(p_{t}^{(h, w, c)})
\label{psedu}
\end{equation}
where $p_{t}^{(h, w, c)} = C_{seg}(E(x_{t}^{(h, w, 3)}))$ refers to the segmentation probability map; $\text{\usefont{U}{bbm}{m}{n}1}$ is a function that returns the input if the condition is true or an empty output otherwise; and $k_{c}$ is the class-balanced weights~\cite{zou2018unsupervised}. 

We then fine-tune the segmentation model $(E+C_{seg})$ with target-domain images $x_{t} \in X_{t}$ and the generated pseudo labels $\hat{y}_{t} \in \hat{Y}_{t}$.
The loss function and training objective of MLCM regularized self-training (MLCMR-ST) is defined as:

\begin{equation}
\min_{\theta_{E}} \min_{\theta_{C_{seg}}} \mathcal{L}^{r}_{ST}(E, C_{seg}) = \sum_{h, w} \sum_{c} -\hat{y}_{t}^{(h, w, c)} \log p_{t}^{(h, w, c)}
\end{equation}
where $\hat{y}_{t}^{(h, w, c)} \in (0, 1)^{(H, W, C)}$ is the one-hot representation transformed from $\hat{y}_{t}^{(h, w)} \in (1, C)^{(H, W)}$; the original self-training (non-regularized) loss $\mathcal{L}_{ST}(E, C_{seg})$ is the same to loss $\mathcal{L}^{r}_{ST}(E, C_{seg})$ except no $\mathcal{M}_{multi}$ regularization in Equation \ref{psedu}.

\section{Experiment}

\subsection{Datasets}
Our experiments are conducted over two synthetic datasets (i.e., GTA5 \cite{richter2016playing} and SYNTHIA \cite{ros2016synthia}) and one real dataset (i.e., Cityscapes \cite{cordts2016cityscapes}). 
GTA5 consists of $24,966$ realistic virtual images with automatically generated pixel-level semantic labels. SYNTHIA contains $9,400$ synthetic images with segmentation labels obtained from a photo-realistic virtual world. Cityscapes is a real-world captured dataset for semantic segmentation. It contains high-resolution images ($2,975/500$ in training/validation set) with human-annotated dense labels.
In our experiments, domain adaptive semantic segmentation networks are trained with the annotated synthetic dataset (GTA5 or SYNTHIA) and the unannotated real dataset (Cityscapes) as in \cite{vu2019advent,guan2020scale,yang2020fda}. The evaluation is performed over the validation set of Cityscapes with the standard mean-Intersection-over-Union (mIoU) metric.

\subsection{Implementation details}
We employ PyTorch toolbox in implementation. All the experiments are conducted on a single Tesla V100 GPU with maximum 12 GB memory usage. Following \cite{tsai2018learning,vu2019advent,luo2019taking,huang2020contextual,guan2020scale}, we adopt Deeplab-V2 architecture \cite{chen2017deeplab} with ResNet-101 pre-trained on ImageNet \cite{deng2009imagenet} as our semantic segmentation backbone $(E + C_{seg})$. 
For multi-level learning, we duplicate and modify $C_{seg}$ to create $C_{region}$ with $N$ output channels as illustrated in Figure \ref{fig:framework}. For a fair comparison to previous works with the VGG backbone, we also apply our networks on VGG-16 \cite{simonyan2014very}. Following \cite{tzeng2014deep}, we use the gradient reverse layer to reverse the entropy based loss between $E$ and ($C_{region}$) during region-level adversarial learning. For domain classifier $C_{image}$, we use a similar structure with \cite{tsai2018learning,vu2019advent}, which includes five 2D-convolution layers with kernel $4 \times 4$ and channel numbers $\{64,128,256,512,1\}$. We use SGD \cite{bottou2010large} to optimize segmentation and local alignment modules ($i.e.$, $E$, $C_{seg}$ and $C_{region}$) with a momentum of $0.9$ and a weight decay of $1e-4$. We optimize the domain classifier $C_{image}$ via Adam with $\beta_{1} = 0.9$, $\beta_{2} = 0.99$. The initial learning rate is set as $2.5e-4$ and decayed by a polynomial policy with a power of $0.9$ as in \cite{chen2017deeplab}. For all experiments, the hyper-parameters $\lambda_{region}$, $\lambda_{ent}$, $\lambda_{image}$ and $N$ are set at $5e-3$, $1e-3$, $1e-3$ and $100$, respectively.

\subsection{Ablation studies}
The proposed multi-level adversarial networks (MLAN) consists of two key components including co-regularized adversarial learning (CR-AL) and regularized adaptation in input and output spaces. We conducted extensive ablation studies to demonstrate the contribution of the two modules to the overall network. Table \ref{tab:abla1} and \ref{tab:abla2} show experimental results.

\textbf{Ablation studies of CR-AL:} We trained 7 segmentation models for the adaptation task GTA5-to-Cityscapes as shown in Table~\ref{tab:abla1}. The 7 network models include 1) \textbf{Baseline} that is trained with \textit{source supervised segmentation loss} $\mathcal{L}_{seg}$ only ($i.e.$, no adaptation), 2) \textbf{RLL}~\cite{huang2020contextual} that is trained using \textit{source supervised reion-level context-relations classification loss} $\mathcal{L}_{region}$ and $\mathcal{L}_{seg}$ only, 3) \textbf{RL-AL}~\cite{huang2020contextual} that is trained using \textit{unsupervised region-level context-relations adaptation loss} $\mathcal{L}_{ent\_region}$, $\mathcal{L}_{region}$ and $\mathcal{L}_{seg}$ only, 4) \textbf{Regularized RL-AL} that is trained using \textit{regularized unsupervised region-level context-relations adaptation loss} $\mathcal{L}^{r}_{ent\_region}$, $\mathcal{L}_{region}$ and $\mathcal{L}_{seg}$ only, 5) \textbf{IL-AL} that is trained using \textit{image-level alignment loss} $\mathcal{L}_{image}$ and $\mathcal{L}_{seg}$ only, 6) \textbf{Regularized IL-AL} that is trained using \textit{regularized image-level alignment loss} $\mathcal{L}^{r}_{image}$ and $\mathcal{L}_{seg}$ only, 7) \textbf{CR-AL} that incorporates and co-regularizes RL-AL and IL-AL via combining \textbf{Regularized RL-AL} and \textbf{Regularized IL-AL}.

\renewcommand\arraystretch{1.2}
\begin{table*}[t]
\caption{Ablation study of co-regularized adversarial learning (CR-AL) on the adaptation from GTA5 dataset to Cityscapes dataset with ResNet-101. RLL denotes the supervised region-level learning on source domain.}
\centering
\begin{scriptsize}
\begin{tabular}{p{3cm}|p{0.5cm}p{0.5cm}p{0.5cm}p{0.5cm}|p{0.5cm}p{0.5cm}|p{1.6cm}}
\hline
\hline
& \multicolumn{4}{c|}{Local Adapt.} & \multicolumn{2}{c|}{Global Adapt.} & \multicolumn{1}{c}{mIoU} 
\\\hline
Method & \multicolumn{1}{c}{$\mathcal{L}_{seg}$} & \multicolumn{1}{c}{$\mathcal{L}_{region}$} &
\multicolumn{1}{c}{$\mathcal{L}_{ent\_region}$} &  \multicolumn{1}{c|}{$\mathcal{L}^{r}_{ent\_region}$} & \multicolumn{1}{c}{$\mathcal{L}_{image}$} &  \multicolumn{1}{c|}{$\mathcal{L}^{r}_{image}$}
\\\hline
Baseline &\multicolumn{1}{c}{\checkmark} & &   &   &  &  &\multicolumn{1}{c}{36.6}\\
RLL~\cite{huang2020contextual} &\multicolumn{1}{c}{\checkmark} &\multicolumn{1}{c}{\checkmark} &   &   &  &  &\multicolumn{1}{c}{39.3}\\
RL-AL~\cite{huang2020contextual}
&\multicolumn{1}{c}{\checkmark} &\multicolumn{1}{c}{\checkmark}   &\multicolumn{1}{c}{\checkmark} &  &   &   &\multicolumn{1}{c}{43.7}\\
Regularized RL-AL &\multicolumn{1}{c}{\checkmark} &\multicolumn{1}{c}{\checkmark}  & &\multicolumn{1}{c|}{\checkmark}   &   &   &\multicolumn{1}{c}{46.1}\\
IL-AL  &\multicolumn{1}{c}{\checkmark} & &  &   &\multicolumn{1}{c}{\checkmark} & &\multicolumn{1}{c}{43.3}\\
Regularized IL-AL
&\multicolumn{1}{c}{\checkmark} &   &  &  & &\multicolumn{1}{c|}{\checkmark}   &\multicolumn{1}{c}{46.9}\\
\textbf{CR-AL} &\multicolumn{1}{c}{\checkmark} &\multicolumn{1}{c}{\checkmark} & &\multicolumn{1}{c|}{\checkmark} &   &\multicolumn{1}{c|}{\checkmark}   &\multicolumn{1}{c}{\textbf{49.1}}\\\hline
\end{tabular}
\end{scriptsize}
\label{tab:abla1}
\end{table*}

We evaluated the 7 models with mIoU and Table~\ref{tab:abla1} shows experimental results. It is shown that the \textbf{Baseline} performs poorly because of the domain discrepancy between synthetic GTA5 and real Cityscapes datasets. \textbf{RLL} outperforms \textbf{Baseline} by $3.3\%$, indicating learning local context-relations on source domain without any adaptation can slightly enhance the adaptation ability of deep segmentation networks. In addition, \textbf{RL-AL} outperforms \textbf{RLL} clearly, which demonstrates the effectiveness of transferring the learnt local context-relations from source domain to target domain ($i.e.$, enforcing the segmentation of unlabelled target images with source-like context-relations via adversarial learning). Further, \textbf{Regularized RL-AL} and \textbf{Regularized IL-AL} outperform \textbf{RL-AL} and \textbf{IL-AL} by a large margin, respectively, demonstrating the importance of mutual regularization in domain adaptive segmentation. Finally, \textbf{CR-AL} performs clearly the best, showing that the region-level adversarial learning (RL-AL) and image-level adversarial learning (IL-AL) are actually complementary to each other.

\textbf{Ablation studies of input and output adaptation:} We trained 6 segmentation networks for the adaptation task GTA5-to-Cityscapes as shown in Table~\ref{tab:abla2}. The 6 models include: 1) \textbf{CR-AL}, 2) \textbf{+IT} that augments target-style source images during CR-AL training based on the \textbf{image translation loss} $\mathcal{L}_{IT}$, 3) \textbf{+RIT} that augments target-style source images during CR-AL training based on the \textit{regularized image translation loss} $\mathcal{L}^{r}_{IT}$, 4) \textbf{+ST} that fine-tunes the trained CR-AL segmentation network by the \textbf{self-training loss} $\mathcal{L}_{ST}$, 5) \textbf{+RST} that fine-tunes the trained CR-AL segmentation network by the \textbf{regularized self-training loss} $\mathcal{L}^{r}_{ST}$, 6) \textbf{MLAN} that incorporates both regularized input and output space domain adaptation.

We evaluated the 6 models with mIoU and Table~\ref{tab:abla1} shows experimental results. It is clear that both \textbf{+IT} and \textbf{+ST} outperform the original CR-AL model, which indicates the CR-AL is complementary to the domain adaptation in input and output spaces. Nevertheless, \textbf{+RIT} and \textbf{+RST} outperform \textbf{+IT} and \textbf{+ST} by a large margin, respectively, demonstrating the importance of introducing multi-level consistencies for both input-space ($e.g.$, image translation) and output-space ($e.g.$, self-training) domain adaptation. Further, \textbf{MLAN} performs clearly the best, which shows that the regularized image translation (RIT) and regularized self-training (RST) are actually complementary to each other.

\renewcommand\arraystretch{1.2}
\begin{table*}[t]
\caption{Ablation study of multi-level adversarial networks (MLAN) that include both co-regularized adversarial learning (CR-AL) and the regularized input and output space domain adaptation (with ResNet-101 as backbone). Experiments are conducted on the task GTA5-to-Cityscapes. IT stands for image translation while RIT stands regularized IT; ST stands for self-training while RST stands for regularized ST.}
\centering
\begin{scriptsize}
\begin{tabular}{p{3cm}|p{1.4cm}p{1.4cm}|p{1.4cm}p{1.4cm}|p{1.6cm}}
\hline
\hline
& \multicolumn{2}{c|}{Input Space Adapt.} & \multicolumn{2}{c|}{Output Space Adapt.} & \multicolumn{1}{c}{mIoU}
\\\hline
Method & \multicolumn{1}{c}{$\mathcal{L}_{IT}$} & \multicolumn{1}{c|}{$\mathcal{L}^{r}_{IT}$} &
\multicolumn{1}{c}{$\mathcal{L}_{ST}$} &  \multicolumn{1}{c|}{$\mathcal{L}^{r}_{ST}$}
\\\hline
CR-AL &   &   &  &  &\multicolumn{1}{c}{49.1}\\
+IT &\multicolumn{1}{c}{\checkmark}   &   &  &  &\multicolumn{1}{c}{50.2}\\
+RIT &   &\multicolumn{1}{c|}{\checkmark}   &  &  &\multicolumn{1}{c}{51.4}\\
+ST &   &   &\multicolumn{1}{c}{\checkmark}  &  &\multicolumn{1}{c}{50.8}\\
+RST &   &   &  &\multicolumn{1}{c|}{\checkmark}  &\multicolumn{1}{c}{52.1}\\
\textbf{MLAN}(+RIT+RST) & &\multicolumn{1}{c|}{\checkmark} &   &\multicolumn{1}{c|}{\checkmark}   &\multicolumn{1}{c}{\textbf{53.6}}\\\hline
\end{tabular}
\end{scriptsize}
\label{tab:abla2}
\end{table*}

\subsection{Comparisons with the State-of-the-Art}
We validate our methods over two challenging synthetic-to-real domain adaptive semantic segmentation tasks ($i.e.$, GTA5-to-Cityscapes and SYNTHIA-to-Cityscapes) with two widely adopted network backbones ResNet-101 and VGG-16. We also compare our network with various state-of-the-art approaches. Tables \ref{tab:bench1} and \ref{tab:bench2} show experimental results. 

\renewcommand\arraystretch{1.1}
\begin{table*}[t]
\caption{Comparisons of the proposed approach with state-of-the-art works over the adaptation from GTA5 \cite{richter2016playing} to Cityscapes\cite{cordts2016cityscapes}.
We highlight the best result in each column in \textbf{bold}.
\textit{'V'} and \textit{'R'} denote the VGG-16 and ResNet-101 backbones, respectively.}
\centering
\begin{tiny}
\begin{tabular}{p{1.6cm}|p{0.1cm}|*{19}{p{0.14cm}}p{0.4cm}}
 \hline
 \hline
 \multicolumn{22}{c}{\textbf{GTA5-to-Cityscapes}} \\[0.05cm]
 \hline
 \hspace{0.5pt}
 Networks &\rot{backbone}    &\rot{road}     &\rot{sidewalk}     &\rot{building}    &\rot{wall}     &\rot{fence}     &\rot{pole}     &\rot{light}     &\rot{sign}     &\rot{veg}     &\rot{terrain}     &\rot{sky}     &\rot{person}     &\rot{rider}     &\rot{car}     &\rot{truck}     &\rot{bus}     &\rot{train}     &\rot{motor}     &\rot{bike}     &mIoU\\
 \hline
 AdaptSeg~\cite{tsai2018learning} &V &87.3 &29.8 &78.6 &21.1 &18.2 &22.5 &21.5 &11.0 &79.7 &29.6 &71.3 &46.8 &6.5 &80.1 &23.0 &26.9 &0.0 &10.6 &0.3 &35.0\\
 AdvEnt~\cite{vu2019advent} &V &86.9 &28.7 &78.7 &28.5 &\textbf{25.2} &17.1 &20.3 &10.9 &80.0 &26.4 &70.2 &47.1 &8.4 &81.5 &26.0 &17.2 &{18.9} &11.7 &1.6 &36.1\\
 CLAN~\cite{luo2019taking} &V &{88.0} &30.6 &79.2 &23.4 &20.5 &26.1 &23.0 &14.8 &{81.6} &{34.5} &72.0 &45.8 &7.9 &80.5 &{\textbf{26.6}} &{29.9} &0.0 &10.7 &0.0 &36.6\\
 CrCDA~\cite{huang2020contextual} &V &86.8	&37.5	&{80.4}	&30.7	&18.1	&26.8	&25.3	&15.1	&81.5	&30.9	&{72.1}	&{52.8}	&{19.0}	&{82.1}	&25.4	&29.2	&10.1	&15.8	&3.7	&{39.1} \\
 BDL~\cite{li2019bidirectional} &V &89.2  &40.9  &81.2  &29.1 &19.2  &14.2  &29.0  &19.6 &83.7  &35.9  &80.7  &54.7  &{\textbf{23.3}}  &82.7 &25.8  &28.0  &2.3 &{\textbf{25.7}} &{19.9} &41.3\\
 FDA~\cite{yang2020fda} &V  &86.1 &35.1 &80.6 &{30.8} &20.4 &27.5 &30.0 &26.0 &82.1 &30.3 &73.6 &52.5 &21.7 &81.7 &24.0 &30.5 &{\textbf{29.9}} &14.6 &{\textbf{24.0}} &42.2\\
 SIM~\cite{wang2020differential} &V &88.1 &35.8 &\textbf{83.1} &25.8 &23.9 &29.2 &28.8 &28.6 &83.0 &\textbf{{36.7}} &\textbf{{82.3}} &53.7 &22.8 &82.3 &26.4 &{\textbf{38.6}} &0.0 &19.6 &17.1 &42.4\\
 SVMin~\cite{guan2020scale} &V &{89.7}	&{42.1}	&82.6	&{29.3}	&22.5	&\textbf{32.3}	&\textbf{35.5}	&\textbf{32.2}	&{84.6}	&35.4	&77.2	&{\textbf{61.6}}	&21.9	&{\textbf{86.2}}	&26.1	&36.7	&7.7	&16.9	&19.4
 &{44.2}\\
 \textbf{Ours} &V &\textbf{90.8}	&\textbf{43.9}	&83.0	&\textbf{32.5}	&24.5	&31.4	&34.2	&31.6	&\textbf{85.4}	&36.1	&81.0	&58.9	&22.8	&86.1	&26.4	&37.8	&24.2	&21.8	&21.3
 &{\textbf{46.0}}\\
 \hline
 AdaptSeg~\cite{tsai2018learning} &R &86.5 &36.0 &79.9 &23.4 &23.3 &23.9 &35.2 &14.8 &83.4 &33.3 &75.6 &58.5 &27.6 &73.7 &32.5 &35.4 &3.9 &30.1 &28.1 &42.4\\
 CLAN~\cite{luo2019taking} &R &87.0 &27.1 &79.6 &27.3 &23.3 &28.3 &35.5 &24.2 &83.6 &27.4 &74.2 &58.6 &28.0 &76.2 &33.1 &36.7 &{6.7} &{31.9} &31.4 &43.2\\
 AdvEnt~\cite{vu2019advent} &R &89.4 &33.1 &81.0 &26.6 &26.8 &27.2 &33.5 &24.7 &{83.9} &{36.7} &78.8 &58.7 &30.5 &{84.8} &38.5 &44.5 &1.7 &31.6 &32.4 &45.5\\
 IDA~\cite{pan2020unsupervised} &R &90.6 &37.1 &82.6 &30.1 &19.1 &29.5 &32.4 &20.6 &85.7 &40.5 &79.7 &58.7 &31.1 &86.3 &31.5 &48.3 &0.0 &30.2 &35.8 &46.3\\
 PatAlign~\cite{tsai2019domain} &R &92.3 &51.9 &82.1 &29.2 &25.1 &24.5 &33.8 &33.0 &82.4 &32.8 &82.2 &58.6 &27.2 &84.3 &33.4 &{46.3} &2.2 &29.5 &32.3 &46.5\\
 CRST~\cite{zou2019confidence} &R &91.0 &{55.4} &80.0 &{33.7} &21.4 &{37.3} &32.9 &24.5 &85.0 &34.1 &80.8 &57.7 &24.6 &84.1 &27.8 &30.1 &{26.9} &26.0 &42.3 &47.1\\
 BDL~\cite{li2019bidirectional} &R &91.0  &44.7  &84.2  &{\textbf{34.6}}  &27.6 &30.2  &36.0  &36.0 &85.0  &{\textbf{43.6}}  &83.0 &58.6 &31.6  &83.3  &35.3  &{49.7} &3.3  &28.8  &35.6 &48.5\\
 CrCDA~\cite{huang2020contextual} &R &{92.4}	&55.3	&{82.3}	&31.2	&{29.1}	&32.5	&33.2	&{35.6}	&83.5	&34.8	&{84.2}	&58.9	&{32.2}	&84.7	&{40.6}	&46.1	&2.1	&31.1	&32.7	&{48.6} \\
 SIM~\cite{wang2020differential} &R &90.6 &44.7 &84.8 &34.3 &28.7 &31.6 &35.0 &37.6 &84.7 &43.3 &85.3 &57.0 &31.5 &83.8 &{\textbf{42.6}} &48.5 &1.9 &30.4 &39.0 &49.2\\
 CAG~\cite{zhang2019category} &R &90.4 &51.6 &83.8 &34.2 &27.8 &38.4 &25.3 &48.4 &85.4 &38.2 &78.1 &58.6 &{\textbf{34.6}} &84.7 &21.9 &42.7 &\textbf{41.1} &29.3 &37.2 &50.2\\
 TIR~\cite{kim2020learning} &R &{92.9} &55.0 &{\textbf{85.3}} &34.2 &{\textbf{31.1}} &34.9 &40.7 &34.0 &85.2 &40.1 &{\textbf{87.1}} &61.0 &31.1 &82.5 &32.3 &42.9 &0.3 &{\textbf{36.4}} &{46.1} &50.2\\
 FDA~\cite{yang2020fda} &R &92.5 &53.3 &82.4 &26.5 &27.6 &36.4 &40.6 &38.9 &82.3 &39.8 &78.0 &62.6 &34.4 &84.9 &34.1 &{\textbf{53.1}} &16.9 &27.7 &{\textbf{46.4}} &50.5\\
 SVMin~\cite{guan2020scale} &R &{92.9}	&{56.2}	&84.3	&34.0	&22.0	&{\textbf{43.1}}	&{\textbf{50.9}}	&{\textbf{48.6}}	&{85.8}	&42.0	&78.9	&{\textbf{66.6}}	&26.9	&{\textbf{88.4}}	&35.2	&46.0	&10.9	&25.4	&39.6 &{51.5}\\
 \textbf{Ours} &R &\textbf{93.6}	&\textbf{58.4}	&84.9	&34.3	&30.6	&42.5	&48.5	&44.6	&\textbf{86.9}	&41.7	&83.4	&61.7	&33.4	&88.0	&41.4	&49.2	&16.2	&31.3	&40.4
 &{\textbf{53.2}}
 \\ \hline
\end{tabular}
\end{tiny}
\label{tab:bench1}
\end{table*}

For the GTA5-to-Cityscapes adaptation, we evaluate the mIoU of 19 classes shared between the source domain and target domain as in \cite{tsai2018learning,vu2019advent,huang2020contextual,guan2020scale}. For a fair comparison in SYNTHIA-to-Cityscapes adaptation, we report both mIoU of 13 classes (mIoU*) and mIoU of 16 classes (mIoU) shared between the source domain and target domain, respectively, as in \cite{tsai2018learning,vu2019advent,huang2020contextual,guan2020scale}.

\renewcommand\arraystretch{1.1}
\begin{table*}[!t]
\caption{Comparisons of the proposed approach with state-of-the-art works over the adaptation task SYNTHIA-to-Cityscapes.
We highlight the best result in each column in \textbf{bold}.
\textit{'V'} and \textit{'R'} denote the VGG-16 and ResNet-101 backbones, respectively. \textit{mIoU} and \textit{mIoU*} stand for the mean intersection-over-union values calculated over sixteen and thirteen categories, respectively.}
\hspace{0.2pt}
\centering
\begin{tiny}
\begin{tabular}{p{1.6cm}|p{0.1cm}|*{16}{p{0.18cm}}p{0.3cm}p{0.4cm}}
 \hline
 \hline
 \multicolumn{20}{c}{\textbf{SYNTHIA-to-Cityscapes}} \\
 \hline
 \hspace{0.5pt}
 Networks &\rot{backbone}     &\rot{road}     &\rot{sidewalk}     &\rot{building}    &\rot{wall}     &\rot{fence}     &\rot{pole}     &\rot{light}     &\rot{sign}     &\rot{veg}    &\rot{sky}     &\rot{person}     &\rot{rider}     &\rot{car}     &\rot{bus}         &\rot{motor}     &\rot{bike}     &mIoU  &mIoU*\\
 \hline
 AdaptSeg~\cite{tsai2018learning} &V &78.9 &29.2 &75.5 &- &- &- &0.1 &4.8 &72.6 &76.7 &43.4 &8.8 &71.1 &16.0 &3.6 &8.4 &- &37.6\\
 AdvEnt~\cite{vu2019advent} &V &67.9 &29.4 &71.9 &6.3 &0.3 &19.9 &0.6 &2.6 &74.9 &74.9 &35.4 &9.6 &67.8 &21.4 &4.1 &15.5 &31.4 &36.6\\
 CLAN~\cite{luo2019taking} &V &{80.4} &{30.7} &74.7 &- &- &- &1.4 &8.0 &77.1 &79.0 &46.5 &8.9 &{73.8} &18.2 &2.2 &9.9 &- &39.3\\
 CrCDA~\cite{huang2020contextual} &V &74.5	&30.5	&{78.6}	&6.6	&{\textbf{0.7}}	&21.2	&2.3	&8.4	&77.4	&79.1	&45.9	&{16.5}	&73.1	&{24.1}	&{9.6}	&14.2	&35.2	&{41.1} \\
 BDL~\cite{li2019bidirectional} &V &72.0 &30.3   &74.5   &0.1   &0.3   &24.6   &10.2 &25.2 &{80.5} &80.0 &54.7   &{\textbf{23.2}}   &72.7   &{24.0} &7.5 &44.9  &39.0 &46.1\\
 FDA~\cite{yang2020fda} &V &{84.2} &{\textbf{35.1}} &{78.0} &6.1 &0.4 &{\textbf{27.0}} &8.5 &22.1 &77.2 &79.6 &55.5 &19.9 &74.8 &{24.9} &{\textbf{14.3}} &40.7 &40.5 &47.3\\
 SVMin~\cite{guan2020scale} &V &{82.5}	&{31.5}	&{77.6}	&{\textbf{7.6}}	&{\textbf{0.7}}	&{26.0}	&{\textbf{12.3}}	&{\textbf{28.4}}	&79.4	&{\textbf{82.1}}	&{58.9}	&21.5	&{82.1}	&22.1	&{9.6}	&{\textbf{49.2}} &{41.9}	&{49.0}\\
\textbf{Ours} &V &\textbf{84.7}	&33.9	&\textbf{78.7}	&7.2	&0.6	&26.5	&11.4	&27.2	&\textbf{81.4}	&\textbf{83.4}	&\textbf{60.2}	&22.7	&81.3	&\textbf{25.8}	&12.6	&47.8
 &{\textbf{42.8}}	&{\textbf{50.1}}\\
 \hline
 PatAlign~\cite{tsai2019domain} &R &82.4 &38.0 &78.6 &8.7 &0.6 &26.0 &3.9 &11.1 &75.5 &84.6 &53.5 &21.6 &71.4 &32.6 &19.3 &31.7 &40.0 &46.5\\
 AdaptSeg~\cite{tsai2018learning} &R &84.3 &42.7 &77.5 &- &- &- &4.7 &7.0 &77.9 &82.5 &54.3 &21.0 &72.3 &32.2 &18.9 &32.3 &- &46.7\\
 CLAN~\cite{luo2019taking} &R &81.3 &37.0 &{80.1} &- &- &- &{16.1} &{13.7} &78.2 &81.5 &53.4 &21.2 &73.0 &32.9 &{22.6} &30.7 &- &47.8\\
 AdvEnt~\cite{vu2019advent} &R &85.6 &42.2 &79.7 &{8.7} &0.4 &25.9 &5.4 &8.1 &{80.4} &84.1 &{57.9} &23.8 &73.3 &36.4 &14.2 &{33.0} &41.2 &48.0\\
 IDA~\cite{pan2020unsupervised} &R &84.3 &37.7 &79.5 &5.3 &0.4 &24.9 &9.2 &8.4 &80.0 &84.1 &57.2 &23.0 &78.0 &38.1 &20.3 &36.5 &41.7 &48.9\\
 TIR~\cite{kim2020learning} &R &{\textbf{92.6}} &{53.2} &79.2 &- &- &- &1.6 &7.5 &78.6 &84.4 &52.6 &20.0 &82.1 &34.8 &14.6 &39.4 &- &49.3\\ 
 CrCDA~\cite{huang2020contextual}  &R &{86.2}	&{44.9}	&79.5	&8.3	&{0.7}	&{27.8}	&9.4	&11.8	&78.6	&{86.5}	&57.2	&{26.1}	&{76.8}	&{39.9}	&21.5	&32.1	&{42.9}	&{50.0}\\
 CRST~\cite{zou2019confidence} &R &67.7 &32.2 &73.9 &10.7 &{\textbf{1.6}} &{\textbf{37.4}} &22.2 &{\textbf{31.2}} &80.8 &80.5 &60.8 &29.1 &82.8 &25.0 &19.4 &45.3 &43.8 &50.1\\
 BDL~\cite{li2019bidirectional} &R &86.0   &46.7   &80.3&-&-&-    &14.1   &11.6 &79.2 &81.3 &54.1   &27.9   &73.7   &{\textbf{42.2}}   &25.7   &45.3  &- &51.4\\
 SIM~\cite{wang2020differential} &R &83.0 &44.0 &80.3 &- &- &- &17.1 &15.8 &80.5 &81.8 &59.9 &{\textbf{33.1}} &70.2 &37.3 &28.5 &45.8 &- &52.1\\
 FDA~\cite{yang2020fda} &R &79.3 &35.0 &73.2 &- &- &- &19.9 &24.0 &61.7 &82.6 &61.4 &31.1 &83.9 &40.8 &{\textbf{38.4}} &{\textbf{51.1}} &- &52.5\\
 CAG~\cite{zhang2019category} &R &84.7 &40.8 &81.7 &7.8 &0.0 &35.1 &13.3 &22.7 &84.5 &77.6 &64.2 &27.8 &80.9 &19.7 &22.7 &48.3 &44.5 &52.6\\
SVMin~\cite{guan2020scale} &R &89.8	&47.7	&{82.3}	&{\textbf{14.4}}	&0.2	&37.1	&{\textbf{35.4}}	&22.1	&{85.1}	&{84.9}	&{\textbf{65.8}}	&25.6	&{\textbf{86.0}}	&30.5	&{31.0}	&{50.7}	&{49.3}	&{56.7}\\
\textbf{Ours} &R &90.6	&\textbf{49.3}	&\textbf{82.6}	&11.6	&1.1	&36.4	&32.9	&23.5	&\textbf{85.8}	&\textbf{86.7}	&63.4	&31.9	&85.1	&40.6	&34.2	&48.9
&{\textbf{50.3}}	&{\textbf{58.1}}\\
\hline
\end{tabular}
\end{tiny}
\label{tab:bench2}
\end{table*}

As shown in Tables \ref{tab:bench1} and \ref{tab:bench2}, the proposed method outperforms all state-of-the-art methods under all four settings, $i.e.$, two adaptation tasks with two backbones. The superior performance of our segmentation model is largely attributed to the introduced multi-level consistencies regularization that
enables concurrent region-level and image-level feature alignment and domain adaptation in both input and output spaces. Without the multi-level consistencies regularization, region-/image-level feature alignment and input/output adaptation tends to lack of image-/region-/multi-level consistencies information and over-minimize their own objectives individually, ultimately leading to sub-optimal segmentation performance in the target domain.

\begin{figure*}[!t]
\centering
\begin{minipage}[h]{0.192\linewidth}
\centering\footnotesize {(a) Image}
\end{minipage}
\vspace{2pt}
\begin{minipage}[h]{0.192\linewidth}
\centering\footnotesize {(b) GT}
\end{minipage}
\begin{minipage}[h]{0.192\linewidth}
\centering\footnotesize {(c) CrCDA~\cite{huang2020contextual}}
\end{minipage}
\begin{minipage}[h]{0.192\linewidth}
\centering\footnotesize {(d) SVMin~\cite{guan2020scale}}
\end{minipage}
\begin{minipage}[h]{0.192\linewidth}
\centering\footnotesize{\textbf{(e) Ours}} 
\end{minipage}
\vspace{6pt}
\centering
\begin{minipage}[h]{0.192\linewidth}
\centering\includegraphics[width=1.0\linewidth]{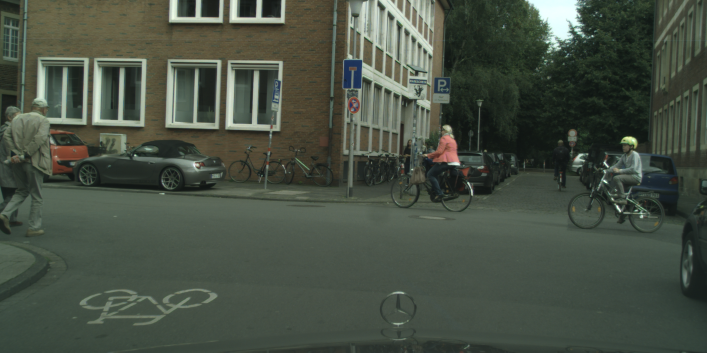}
\end{minipage}
\begin{minipage}[h]{0.192\linewidth}
\centering\includegraphics[width=1.0\linewidth]{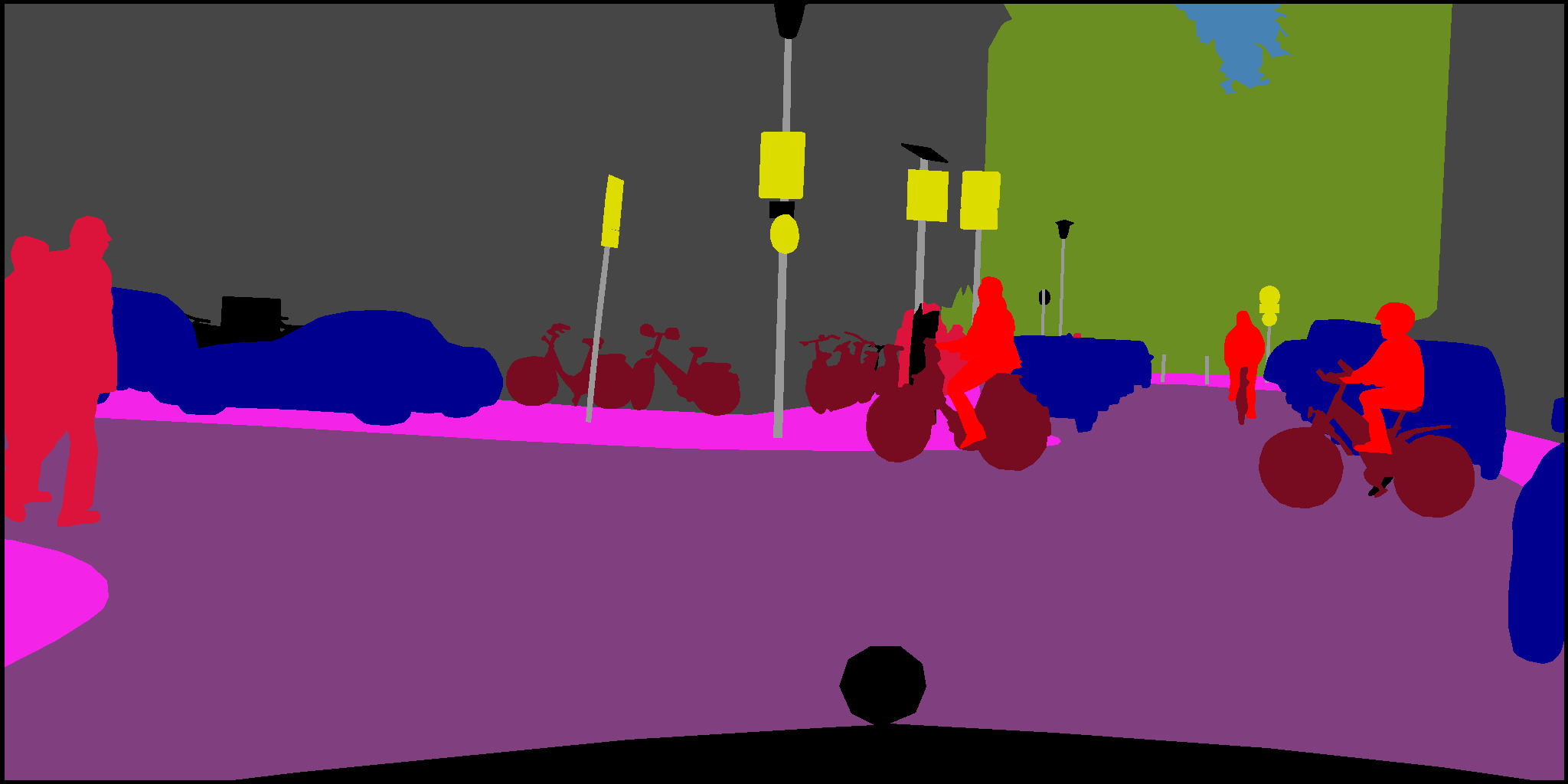}
\end{minipage}
\begin{minipage}[h]{0.192\linewidth}
\centering\includegraphics[width=1.0\linewidth]{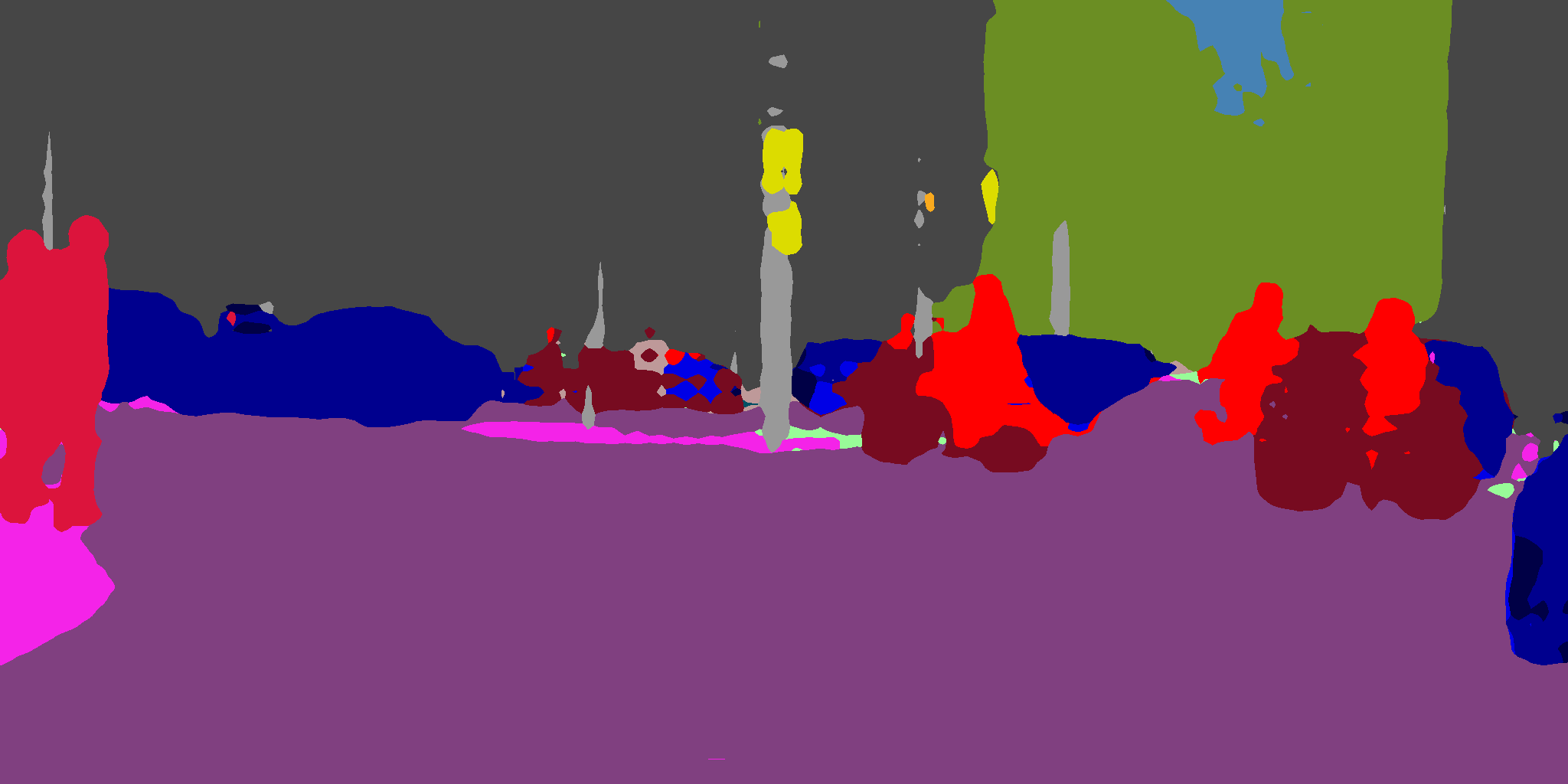}
\end{minipage}
\begin{minipage}[h]{0.192\linewidth}
\centering\includegraphics[width=1.0\linewidth]{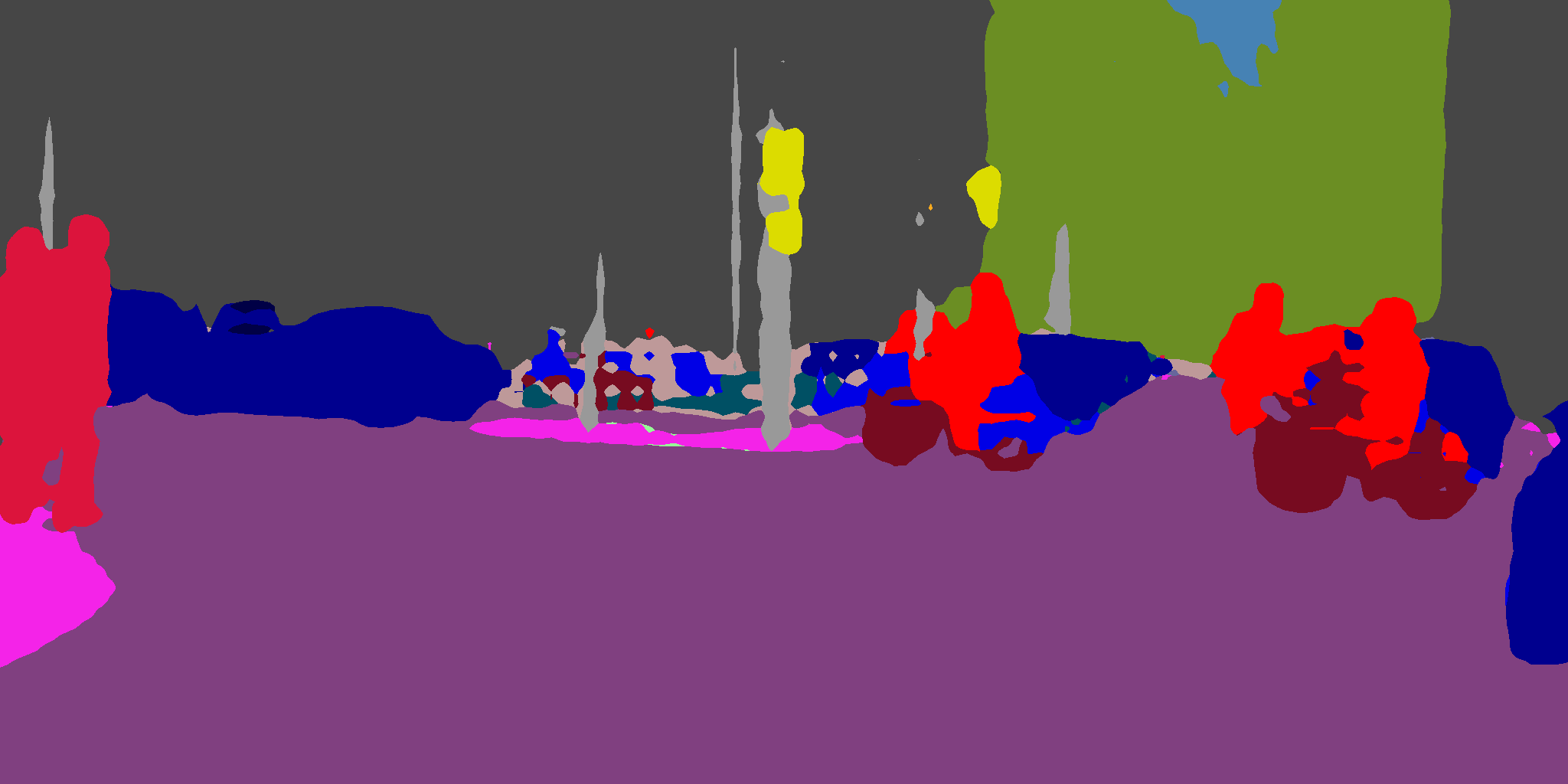}
\end{minipage}
\begin{minipage}[h]{0.192\linewidth}
\centering\includegraphics[width=1.0\linewidth]{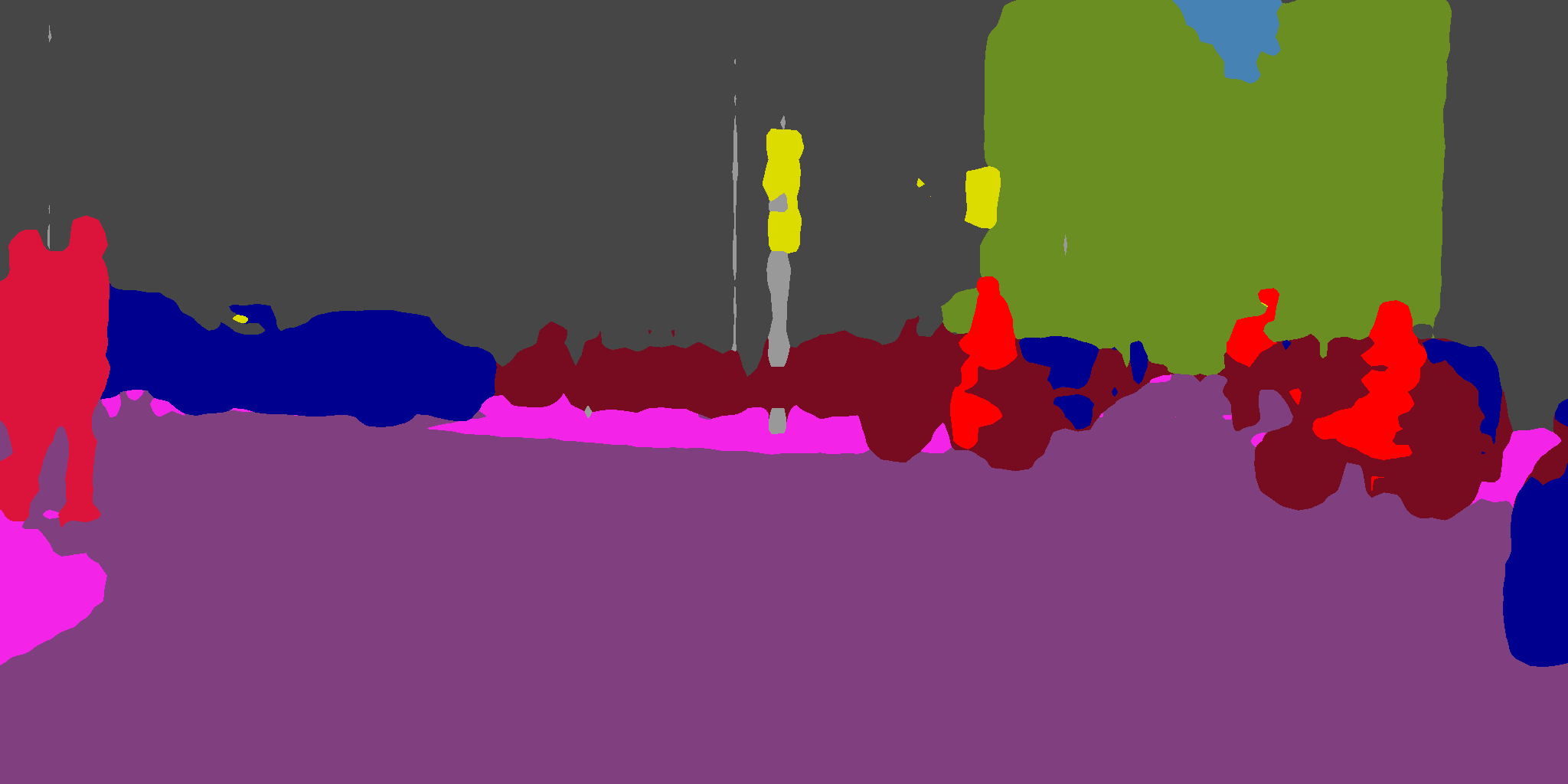}
\end{minipage}
\vspace{6pt}
\centering
\begin{minipage}[h]{0.192\linewidth}
\centering\includegraphics[width=1.0\linewidth]{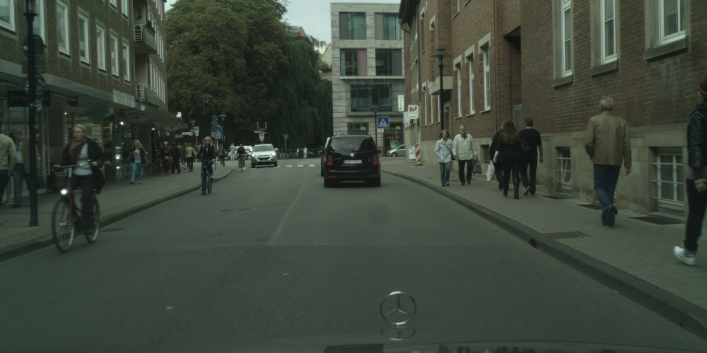}
\end{minipage}
\begin{minipage}[h]{0.192\linewidth}
\centering\includegraphics[width=1.0\linewidth]{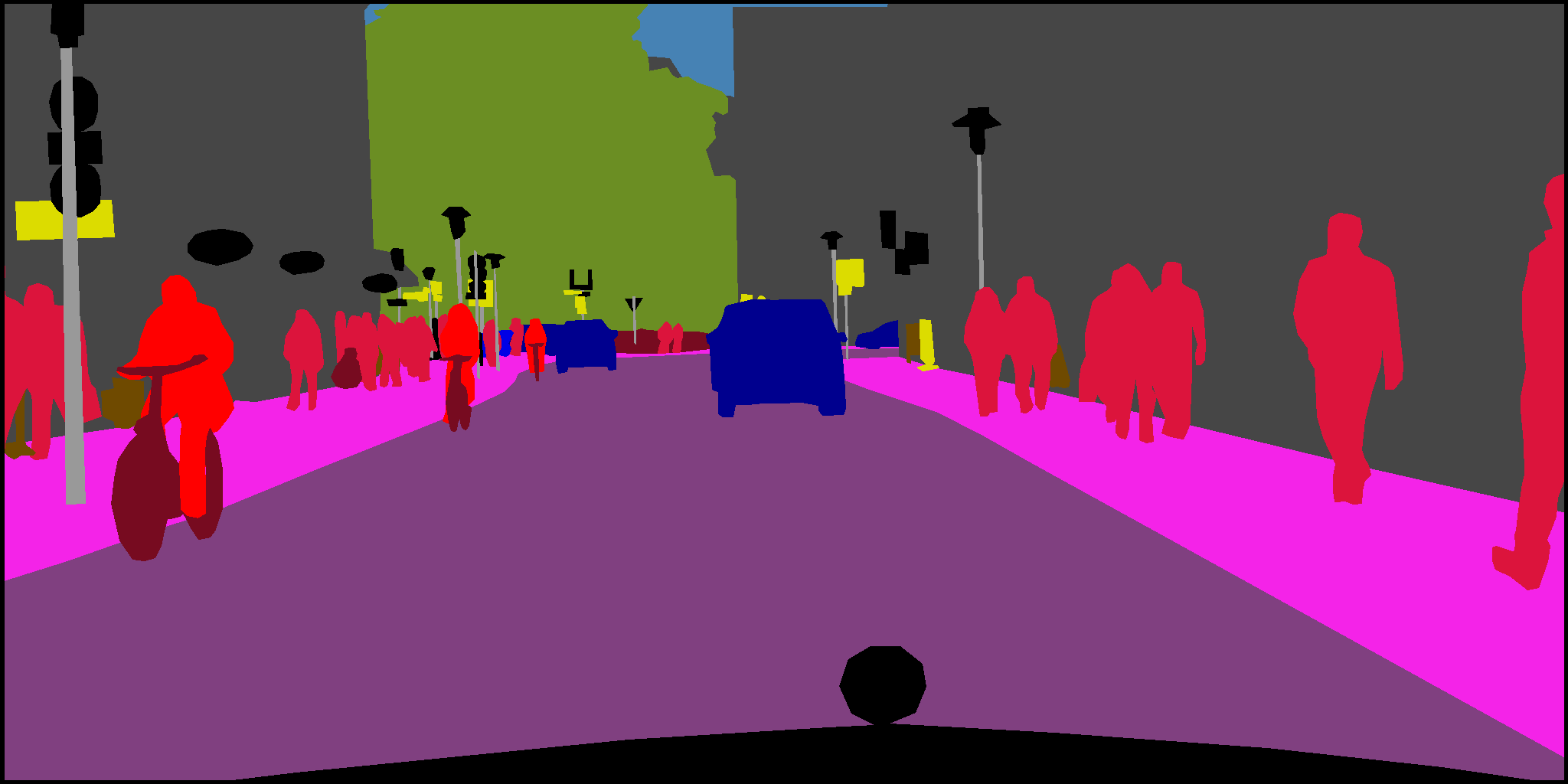}
\end{minipage}
\begin{minipage}[h]{0.192\linewidth}
\centering\includegraphics[width=1.0\linewidth]{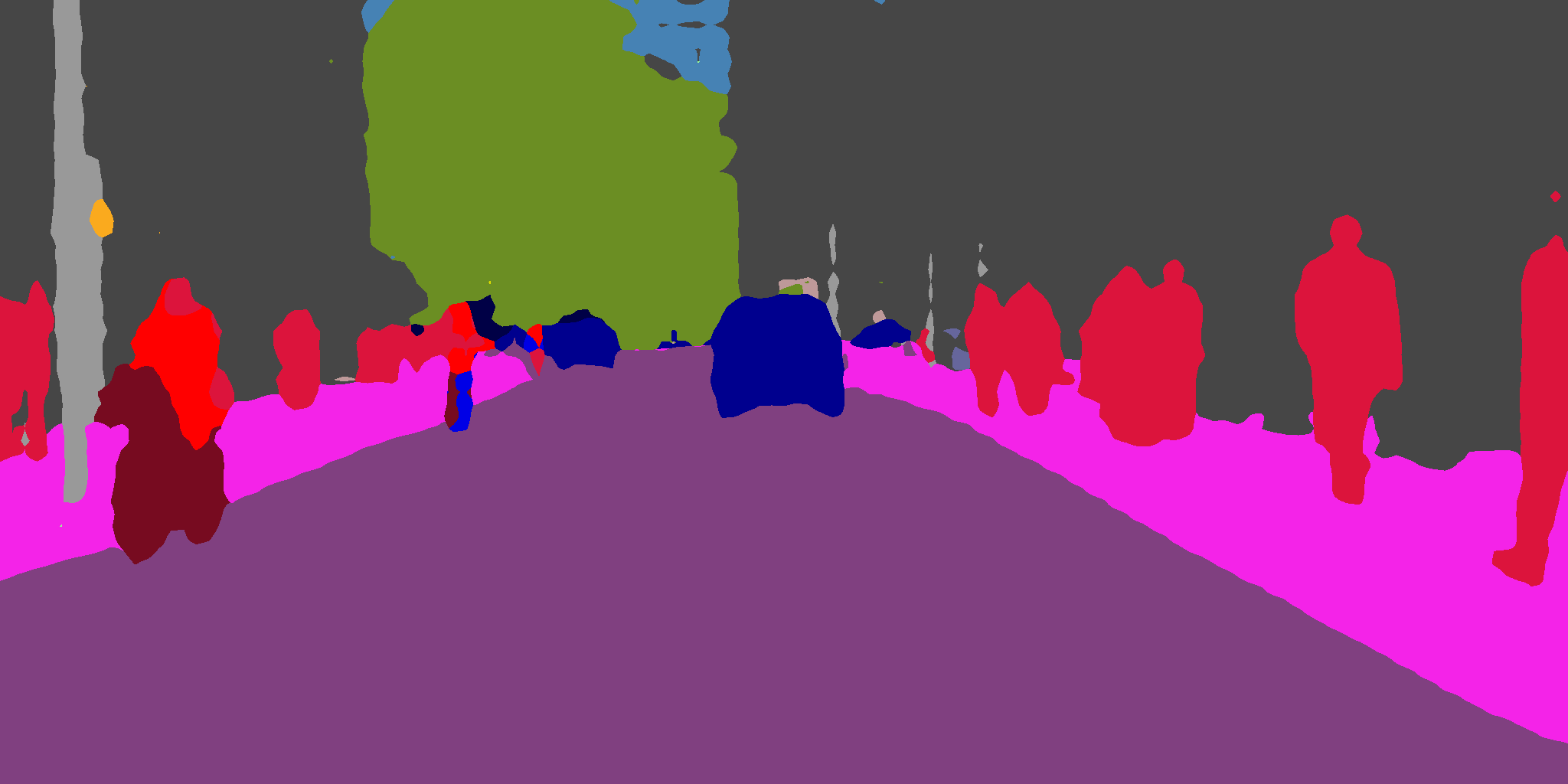}
\end{minipage}
\begin{minipage}[h]{0.192\linewidth}
\centering\includegraphics[width=1.0\linewidth]{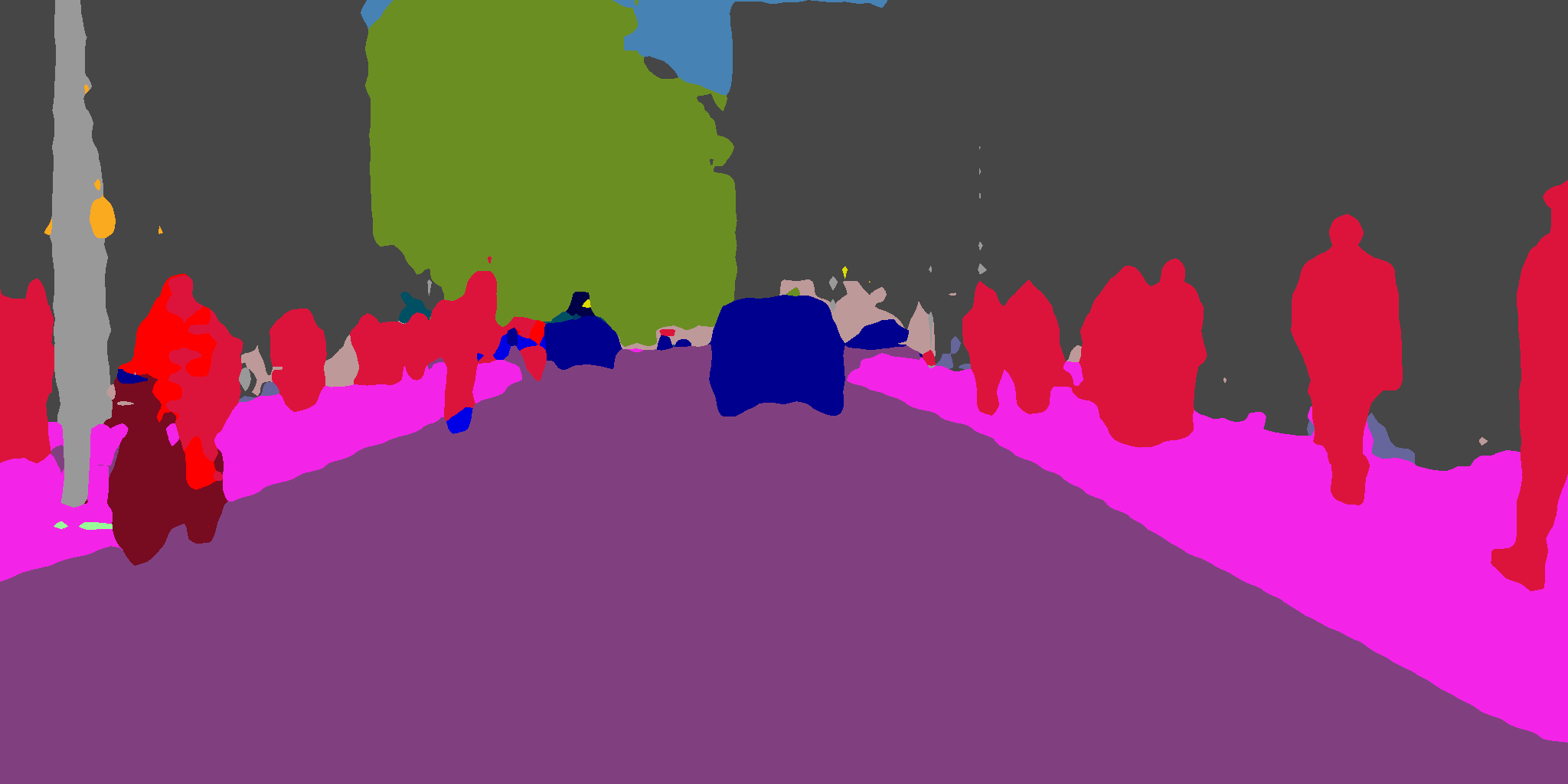}
\end{minipage}
\begin{minipage}[h]{0.192\linewidth}
\centering\includegraphics[width=1.0\linewidth]{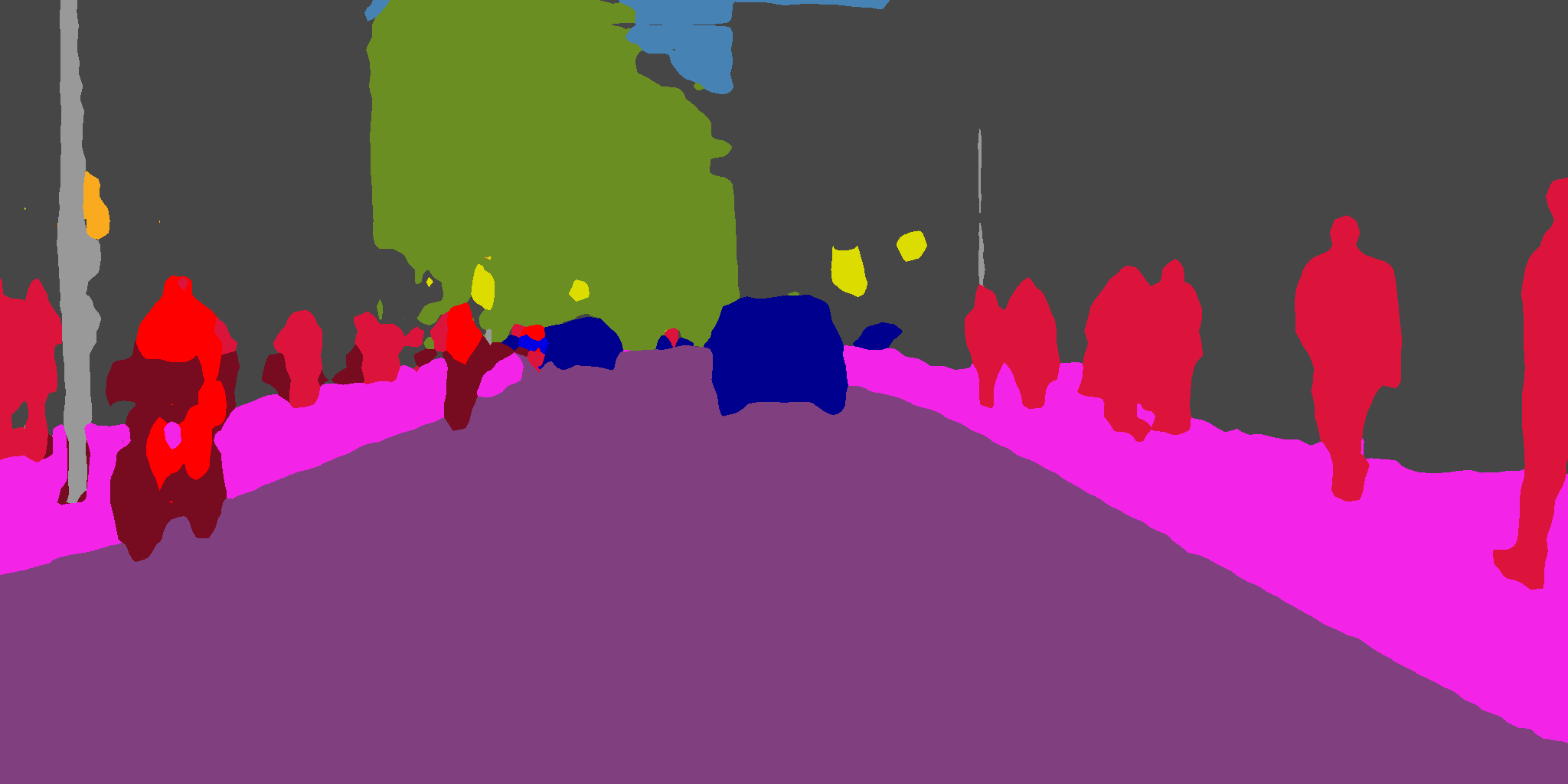}
\end{minipage}
\vspace{6pt}
\centering
\begin{minipage}[h]{0.192\linewidth}
\centering\includegraphics[width=1.0\linewidth]{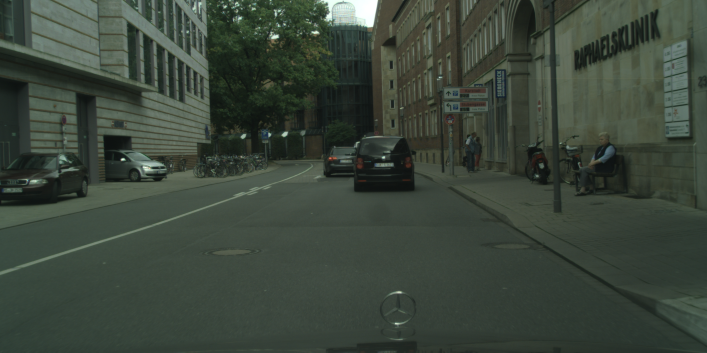}
\end{minipage}
\begin{minipage}[h]{0.192\linewidth}
\centering\includegraphics[width=1.0\linewidth]{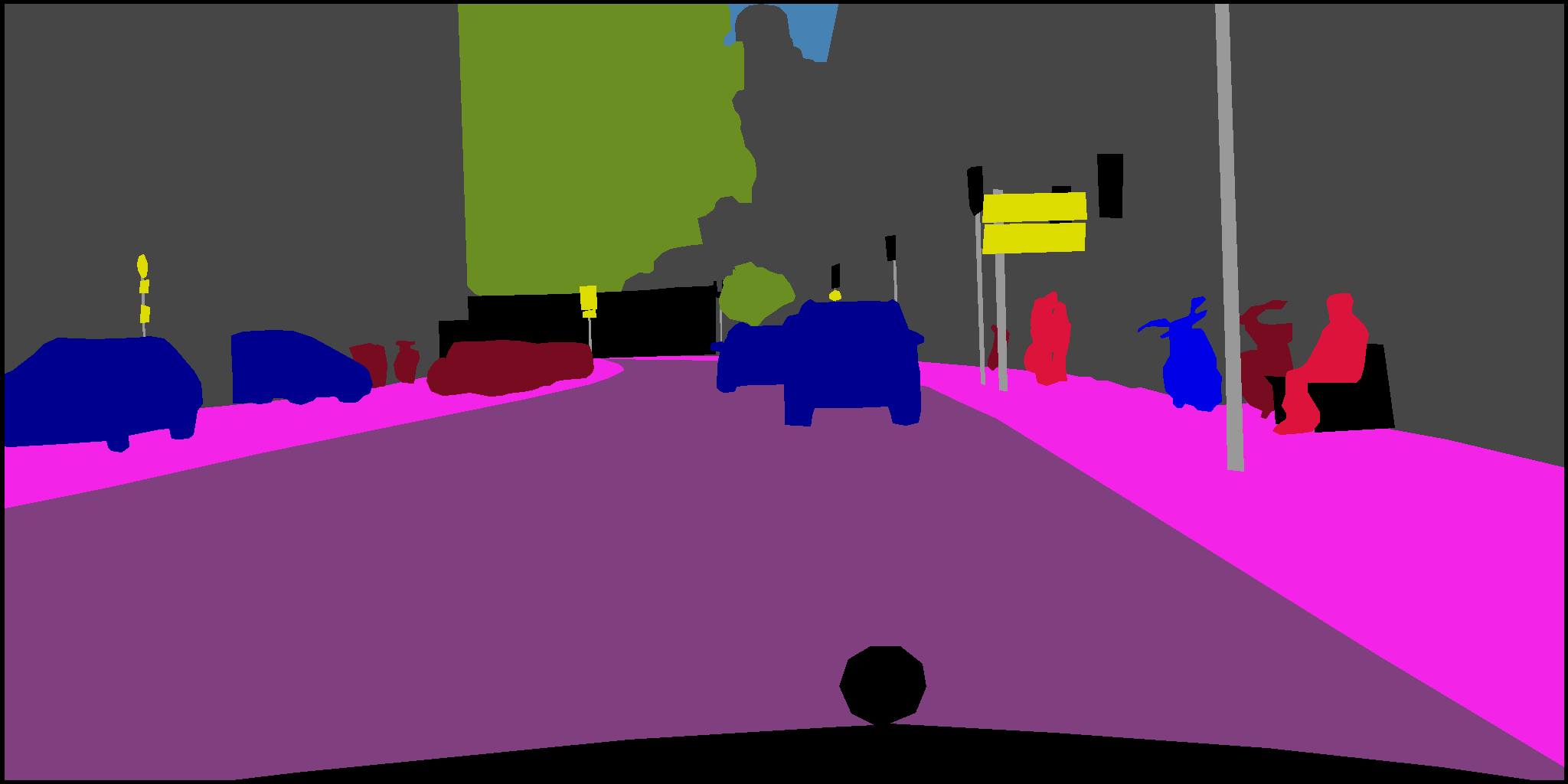}
\end{minipage}
\begin{minipage}[h]{0.192\linewidth}
\centering\includegraphics[width=1.0\linewidth]{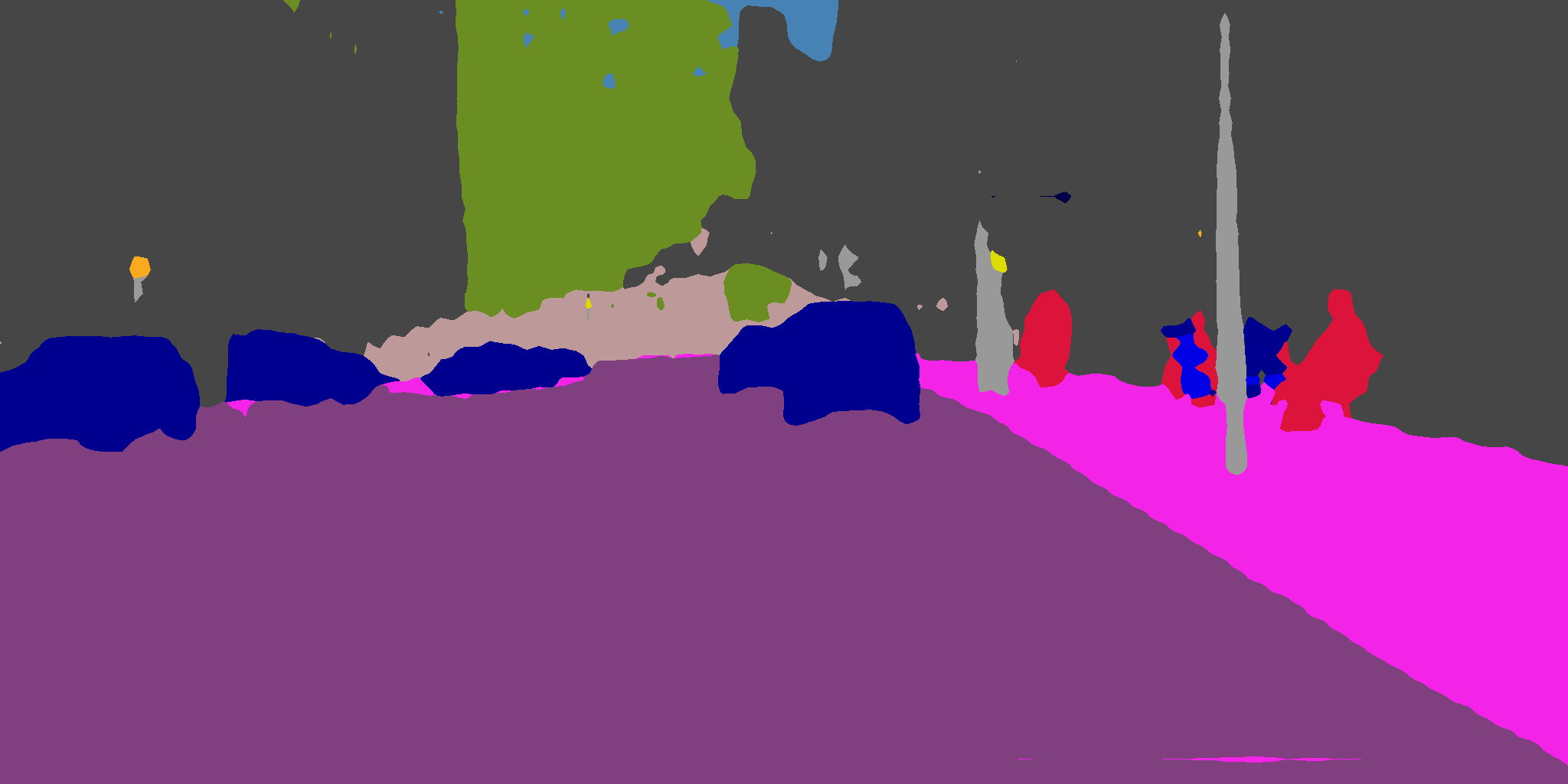}
\end{minipage}
\begin{minipage}[h]{0.192\linewidth}
\centering\includegraphics[width=1.0\linewidth]{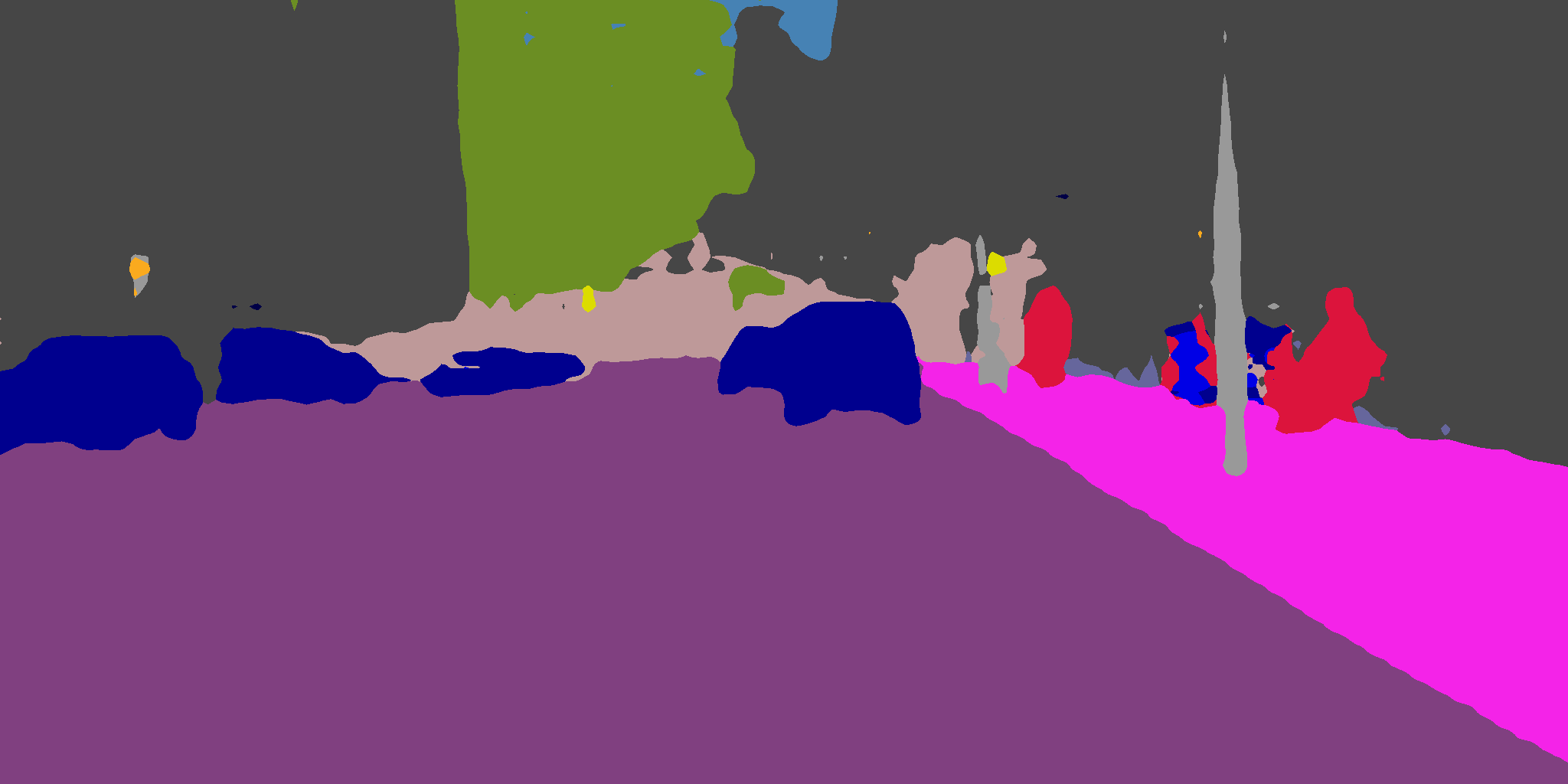}
\end{minipage}
\begin{minipage}[h]{0.192\linewidth}
\centering\includegraphics[width=1.0\linewidth]{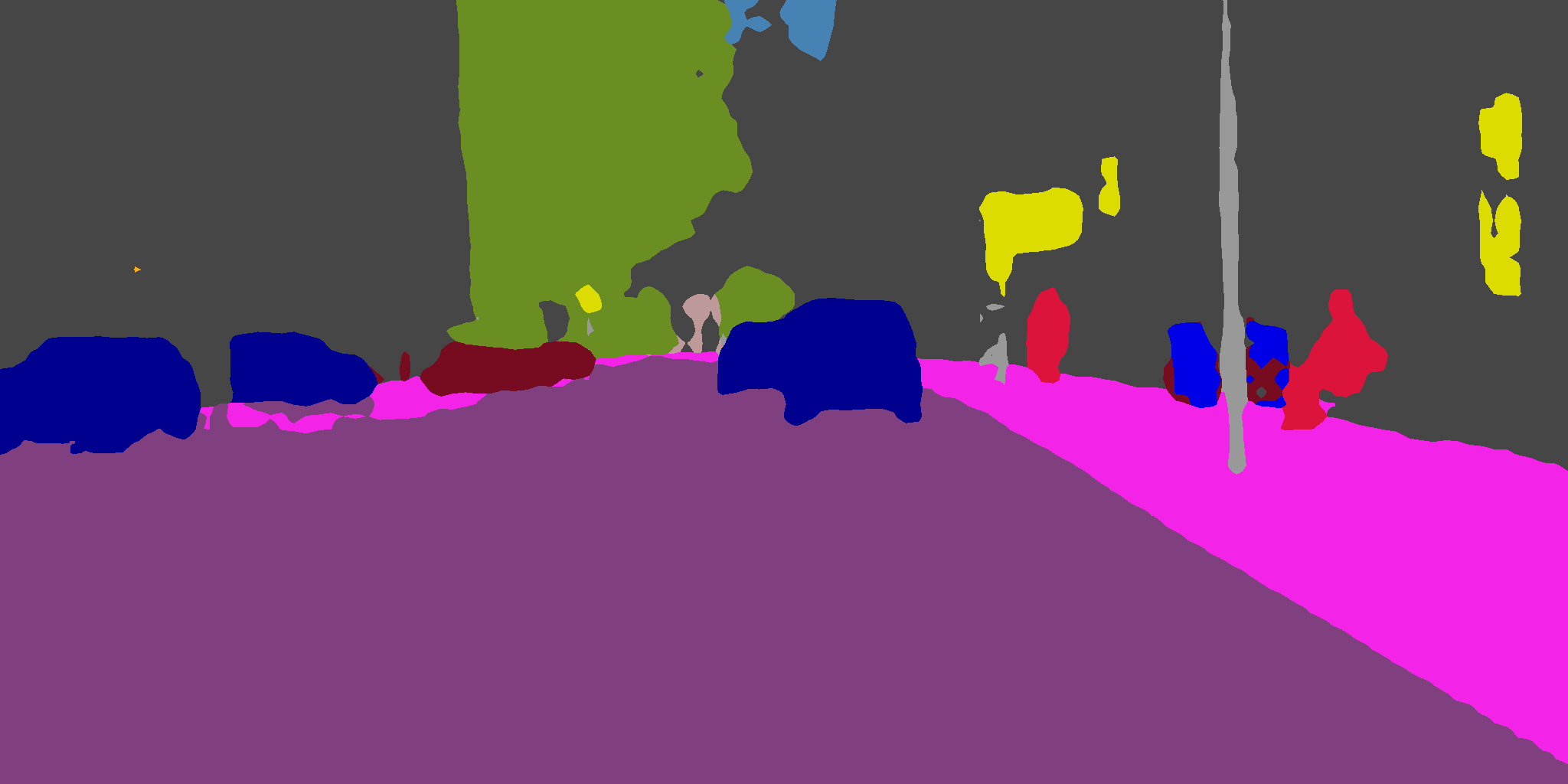}
\end{minipage}
\vspace{6pt}
\centering
\begin{minipage}[h]{0.192\linewidth}
\centering\includegraphics[width=1.0\linewidth]{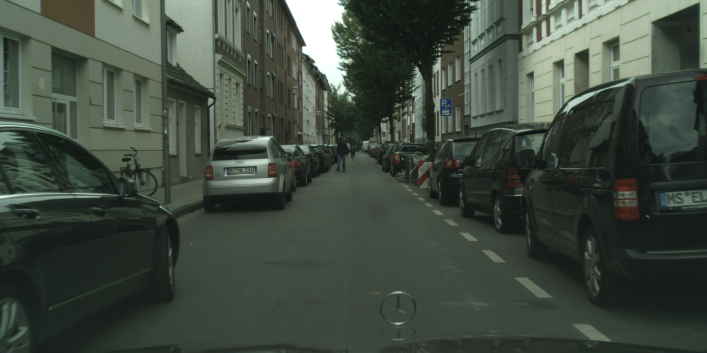}
\end{minipage}
\begin{minipage}[h]{0.192\linewidth}
\centering\includegraphics[width=1.0\linewidth]{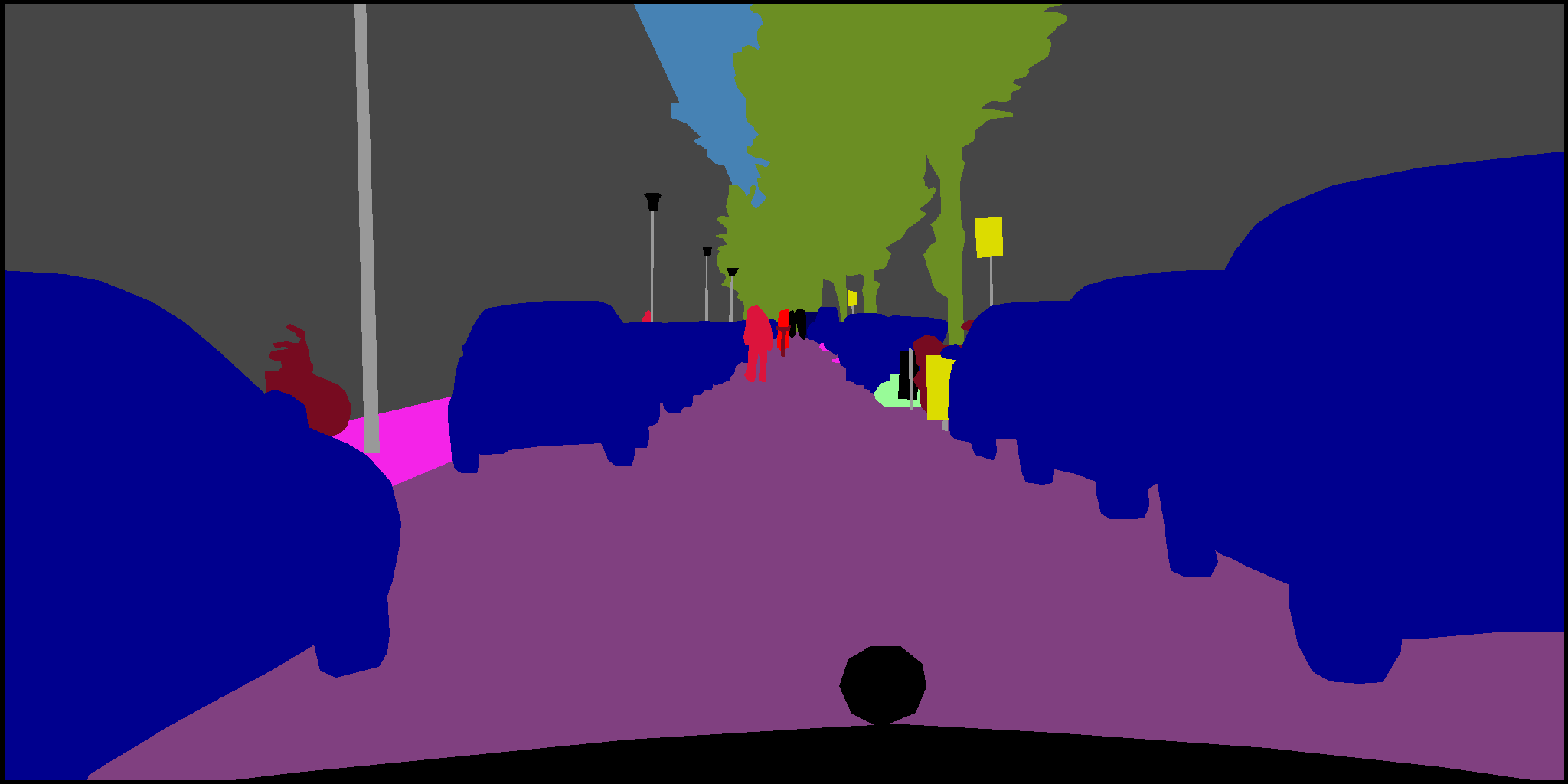}
\end{minipage}
\begin{minipage}[h]{0.192\linewidth}
\centering\includegraphics[width=1.0\linewidth]{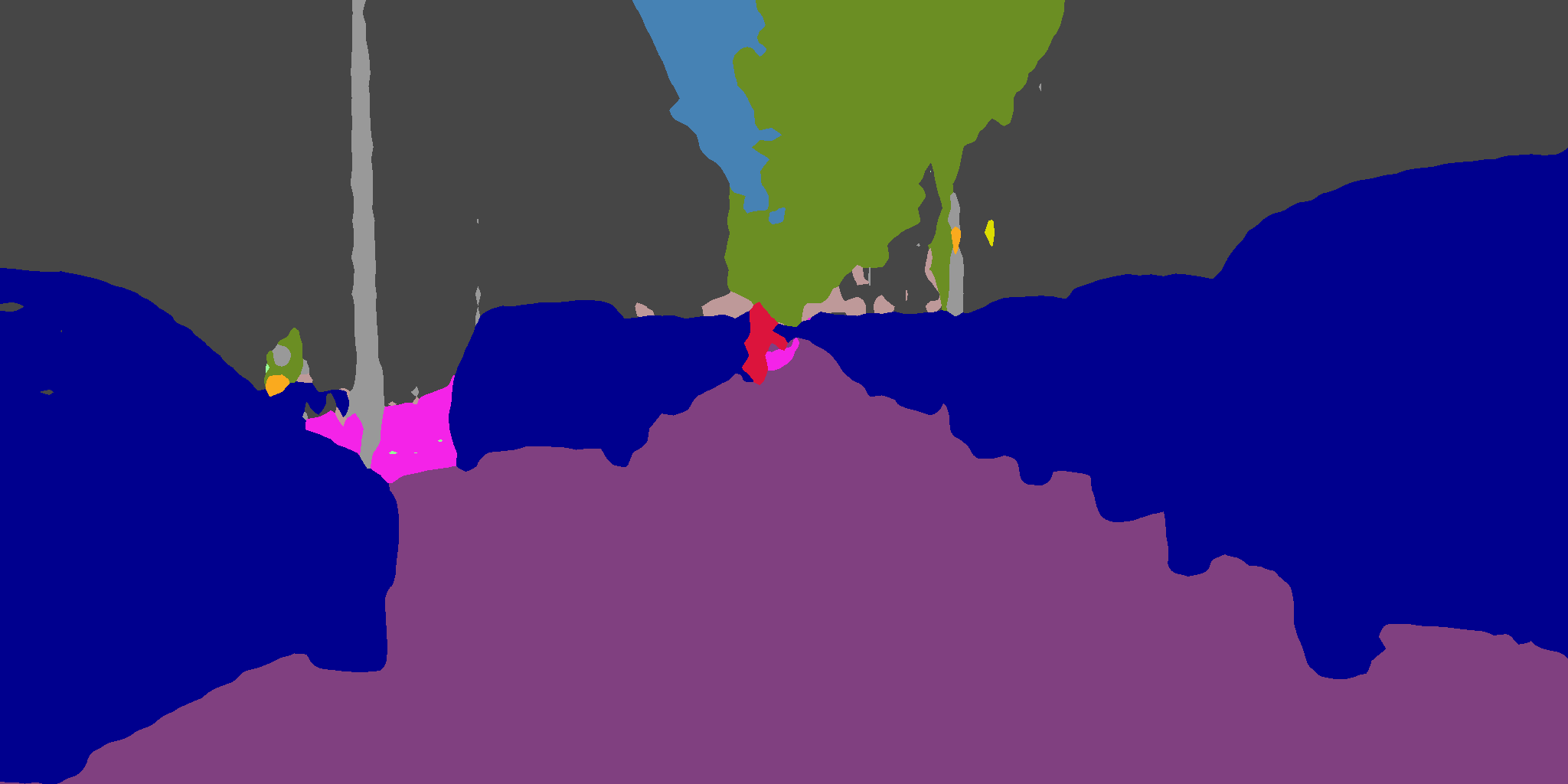}
\end{minipage}
\begin{minipage}[h]{0.192\linewidth}
\centering\includegraphics[width=1.0\linewidth]{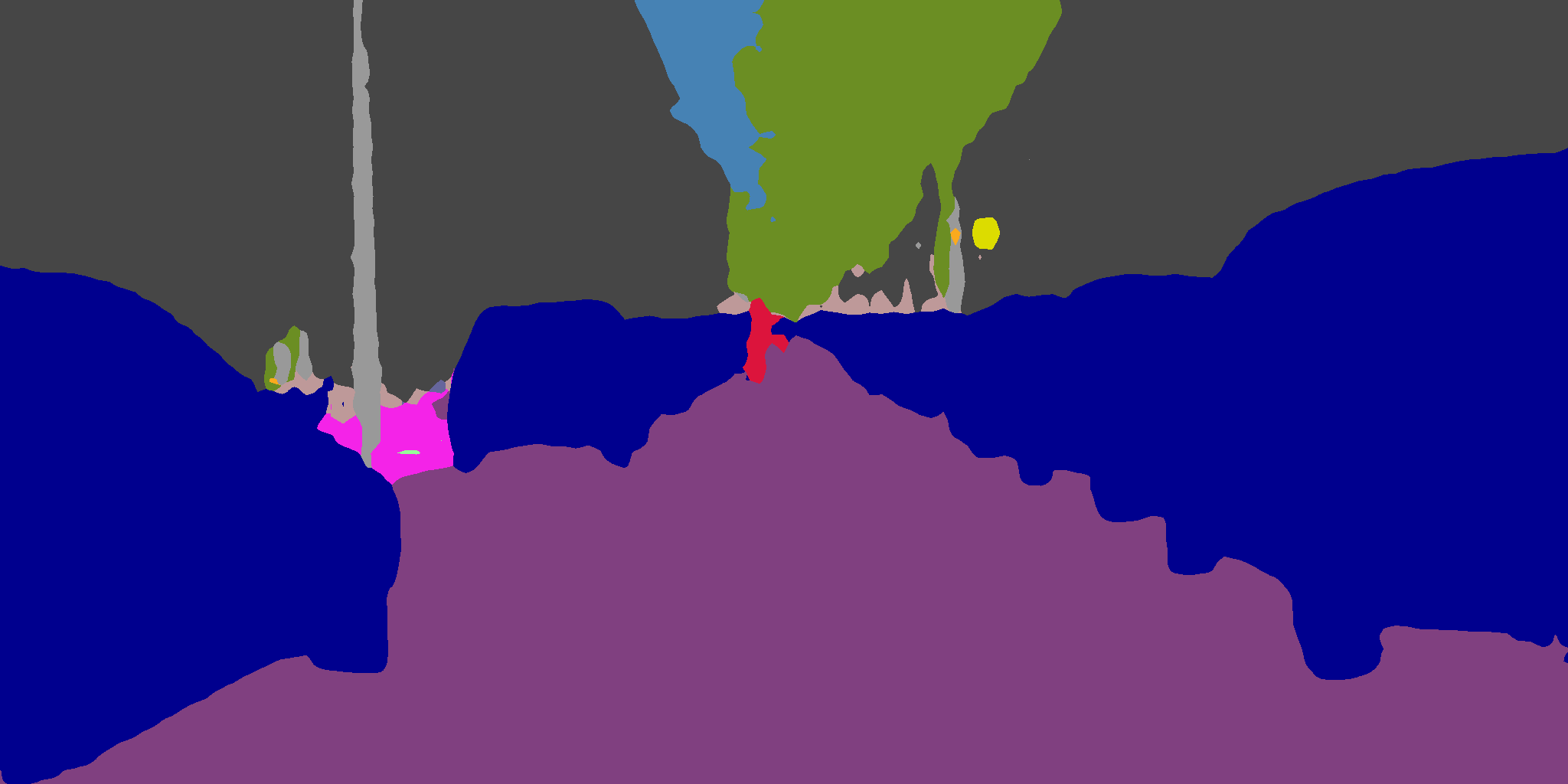}
\end{minipage}
\begin{minipage}[h]{0.192\linewidth}
\centering\includegraphics[width=1.0\linewidth]{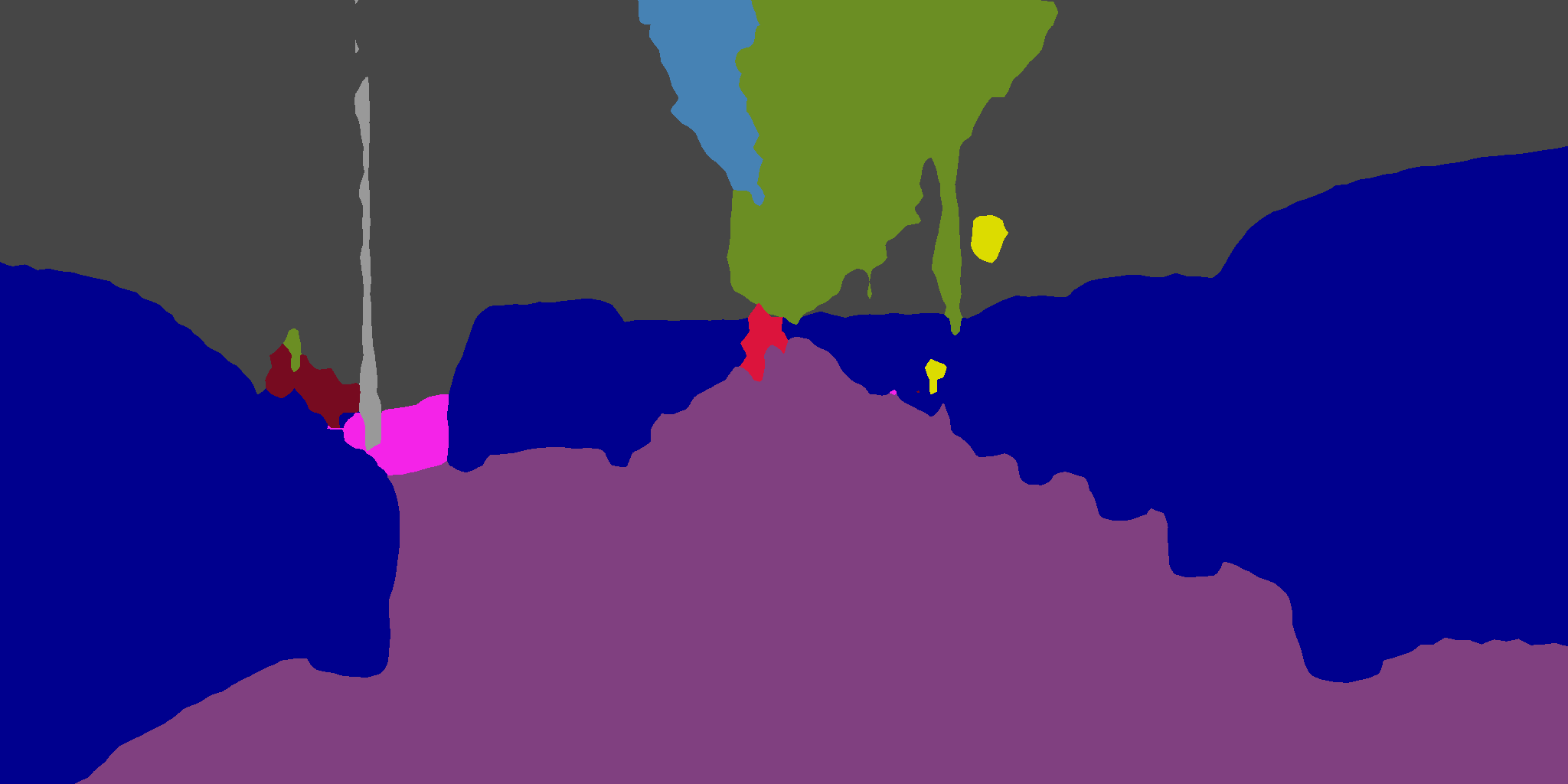}
\end{minipage}
\vspace{6pt}
\centering
\begin{minipage}[h]{0.192\linewidth}
\centering\includegraphics[width=1.0\linewidth]{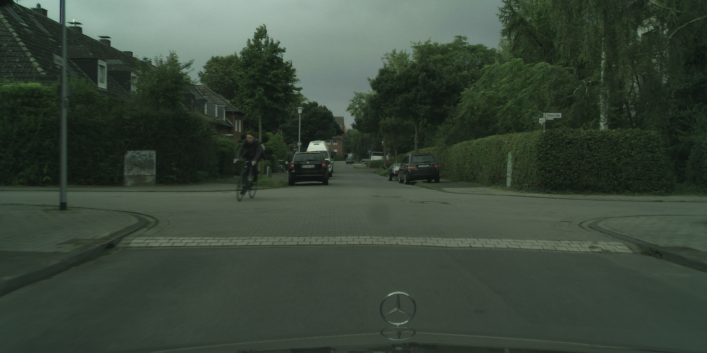}
\end{minipage}
\begin{minipage}[h]{0.192\linewidth}
\centering\includegraphics[width=1.0\linewidth]{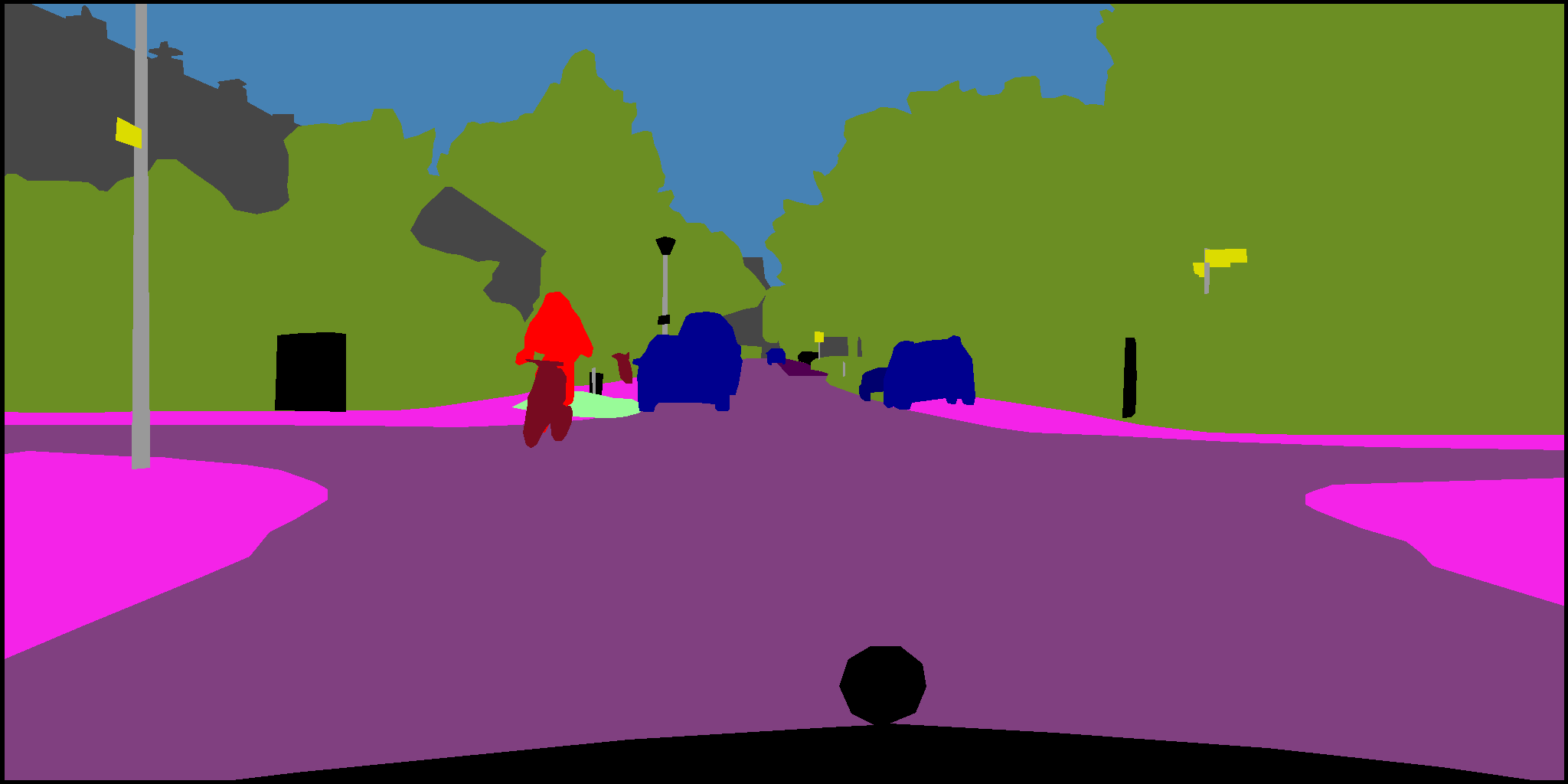}
\end{minipage}
\begin{minipage}[h]{0.192\linewidth}
\centering\includegraphics[width=1.0\linewidth]{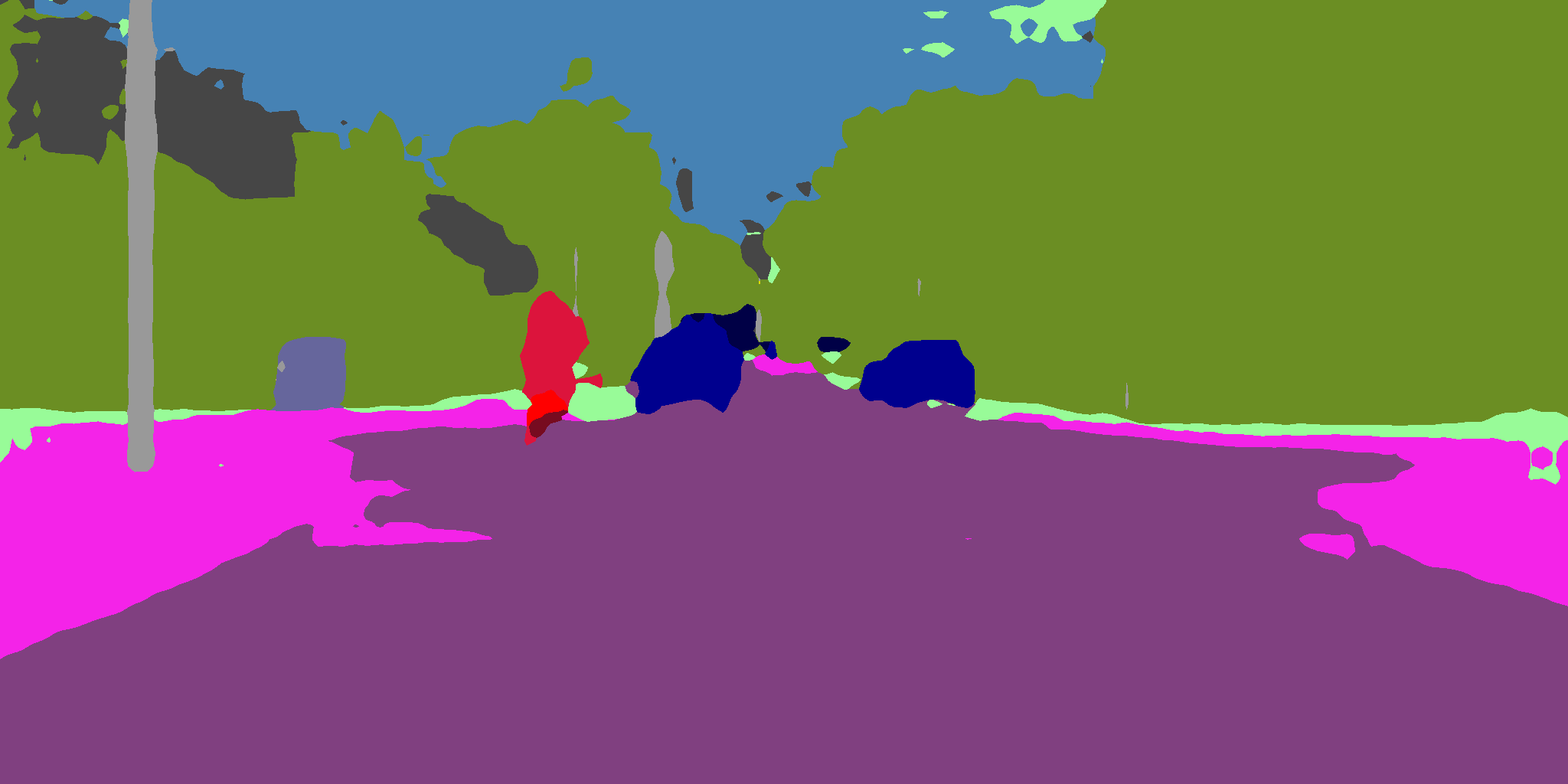}
\end{minipage}
\begin{minipage}[h]{0.192\linewidth}
\centering\includegraphics[width=1.0\linewidth]{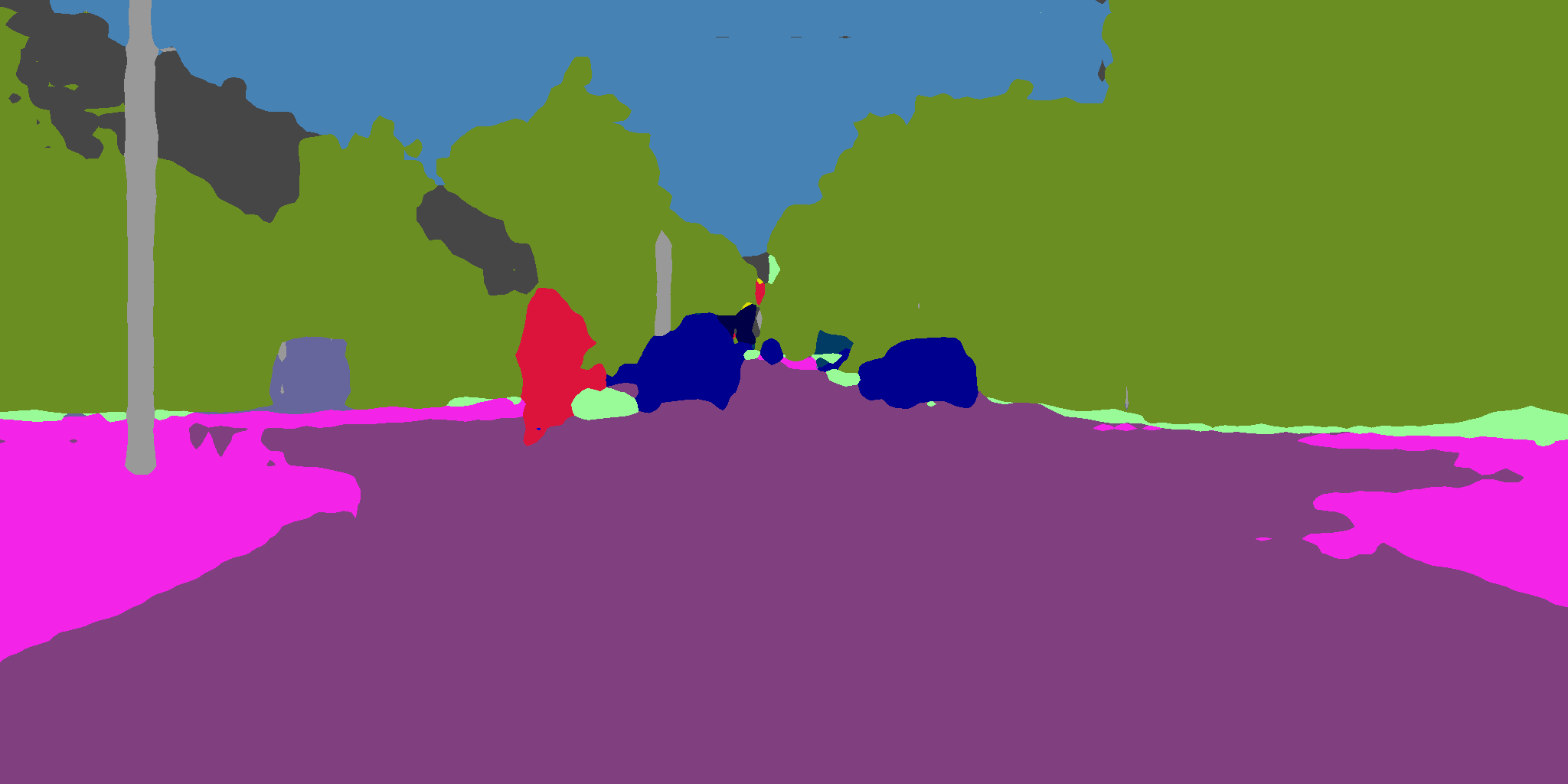}
\end{minipage}
\begin{minipage}[h]{0.192\linewidth}
\centering\includegraphics[width=1.0\linewidth]{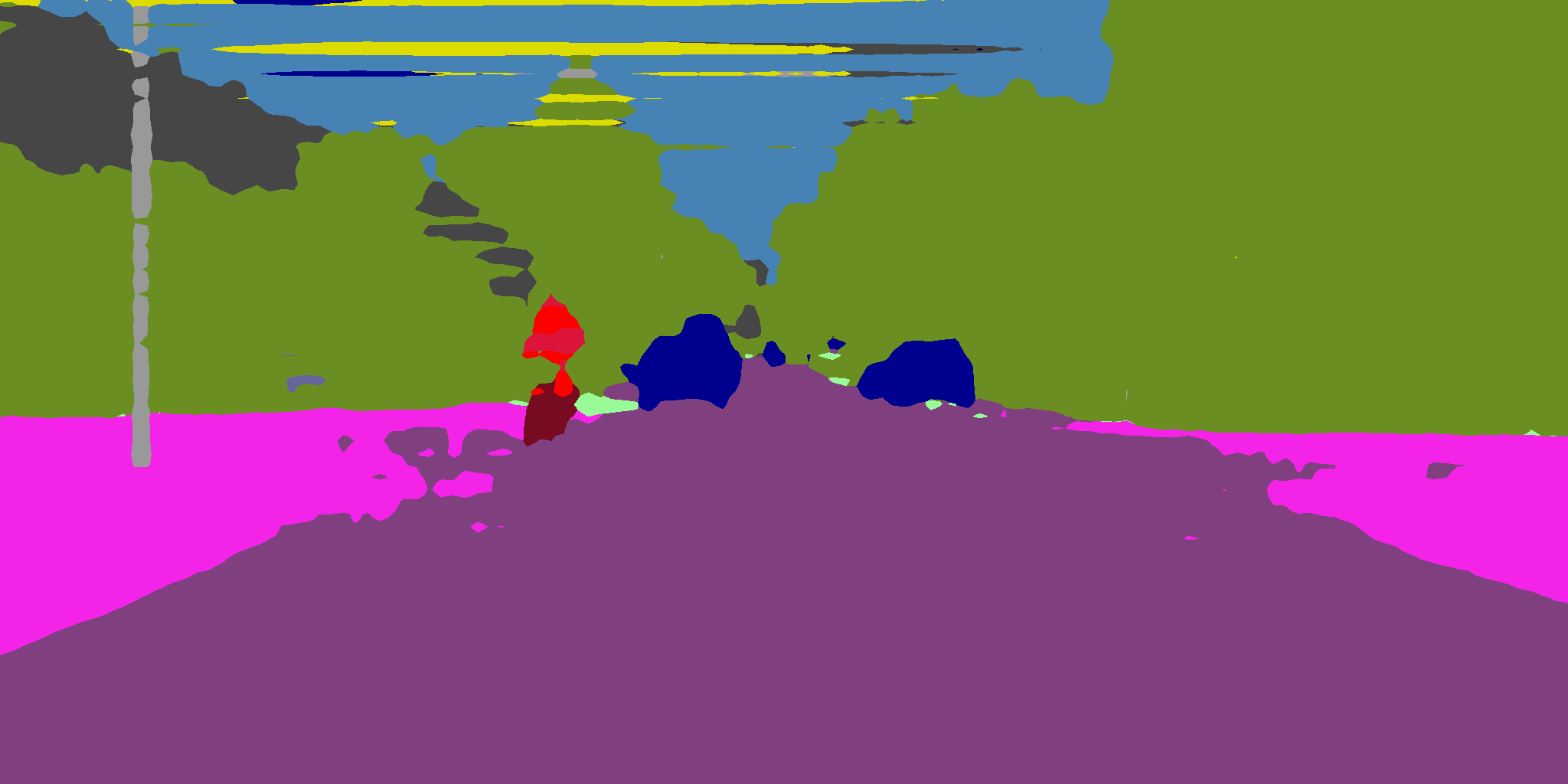}
\end{minipage}
\vspace{6pt}
\centering
\begin{minipage}[h]{0.192\linewidth}
\centering\includegraphics[width=1.0\linewidth]{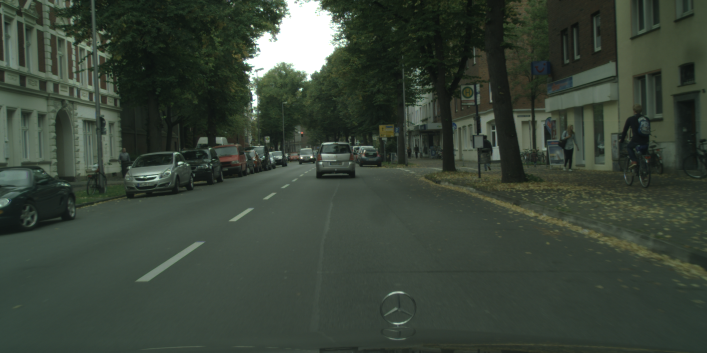}
\end{minipage}
\begin{minipage}[h]{0.192\linewidth}
\centering\includegraphics[width=1.0\linewidth]{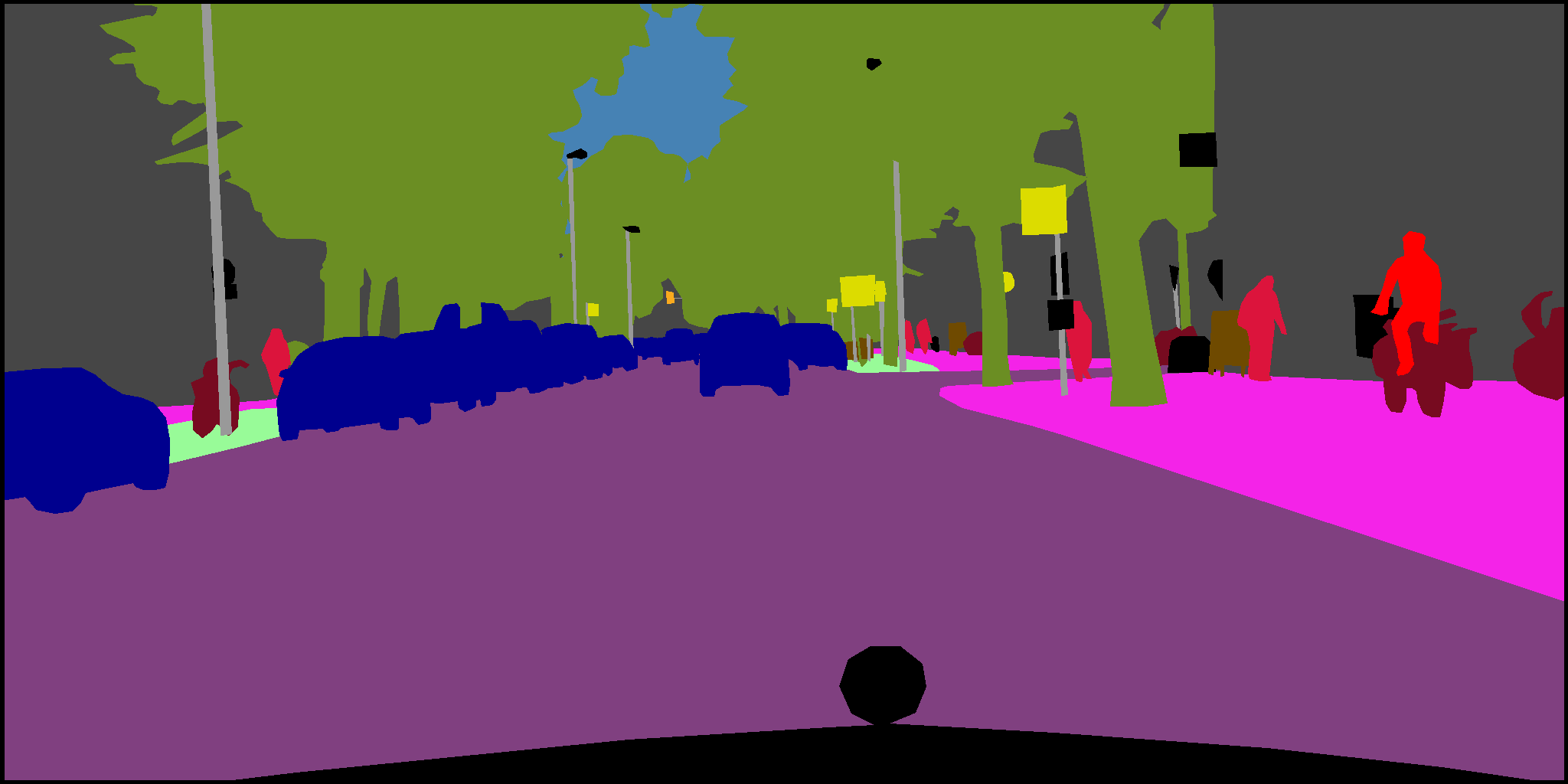}
\end{minipage}
\begin{minipage}[h]{0.192\linewidth}
\centering\includegraphics[width=1.0\linewidth]{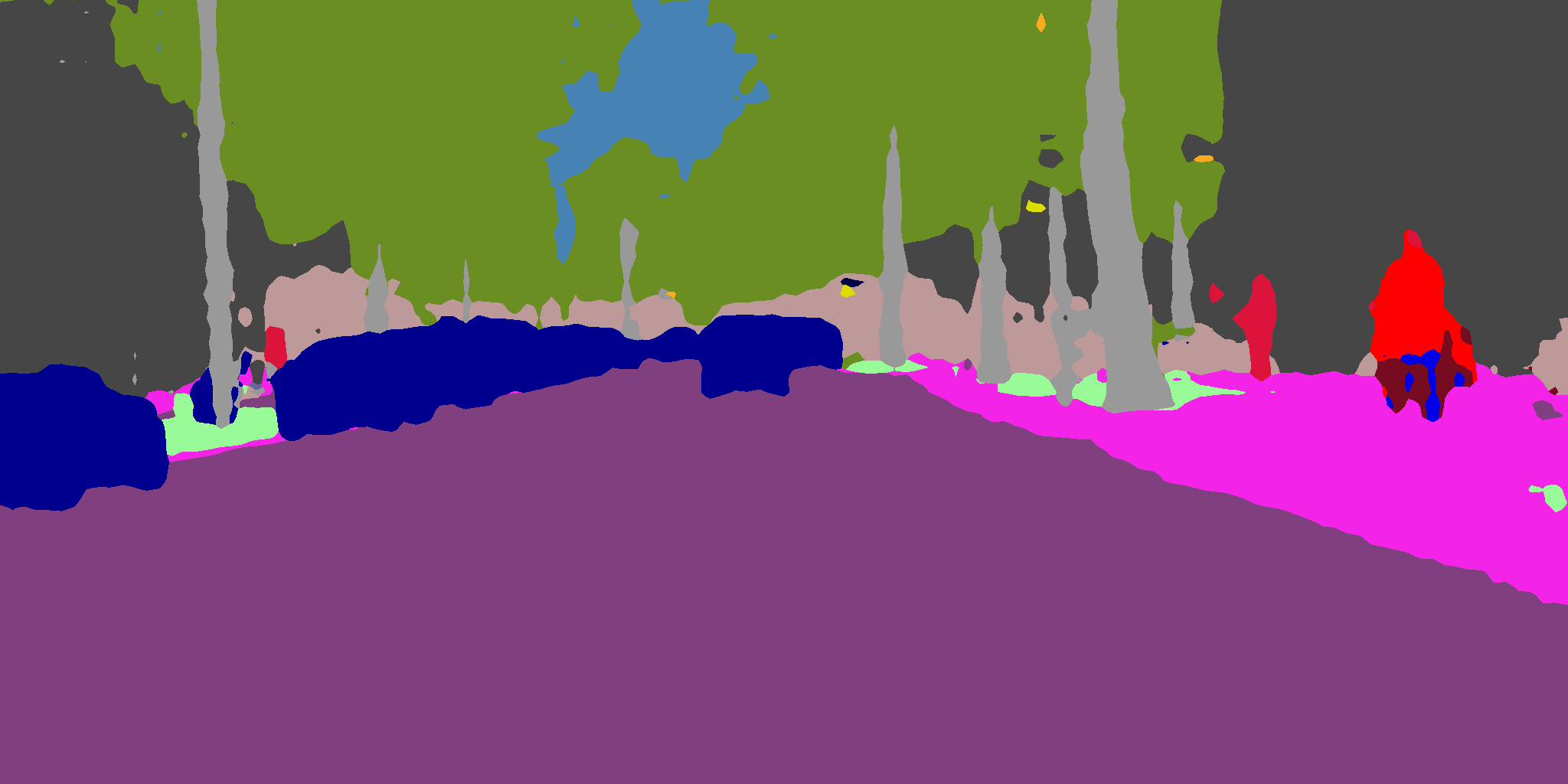}
\end{minipage}
\begin{minipage}[h]{0.192\linewidth}
\centering\includegraphics[width=1.0\linewidth]{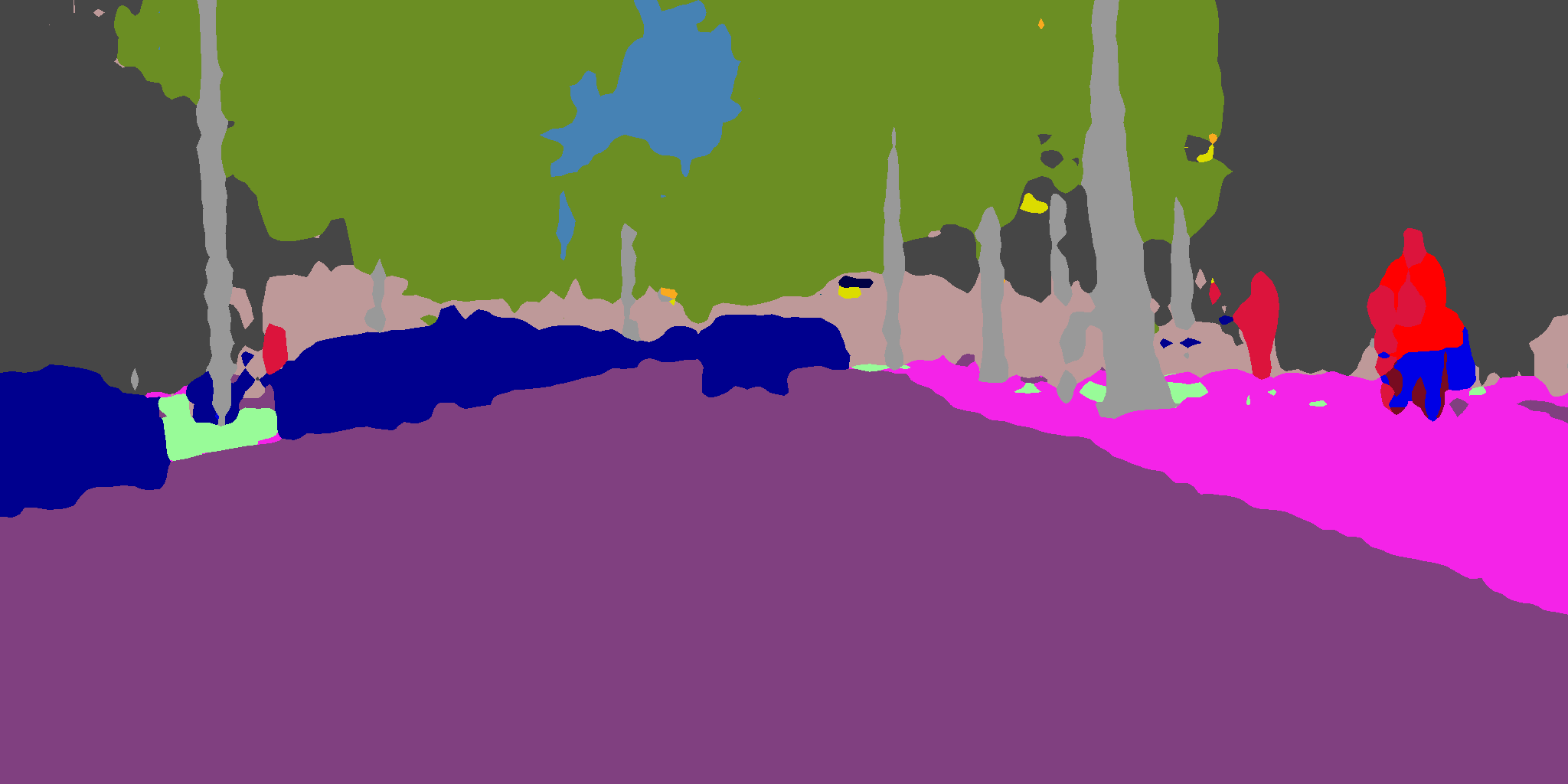}
\end{minipage}
\begin{minipage}[h]{0.192\linewidth}
\centering\includegraphics[width=1.0\linewidth]{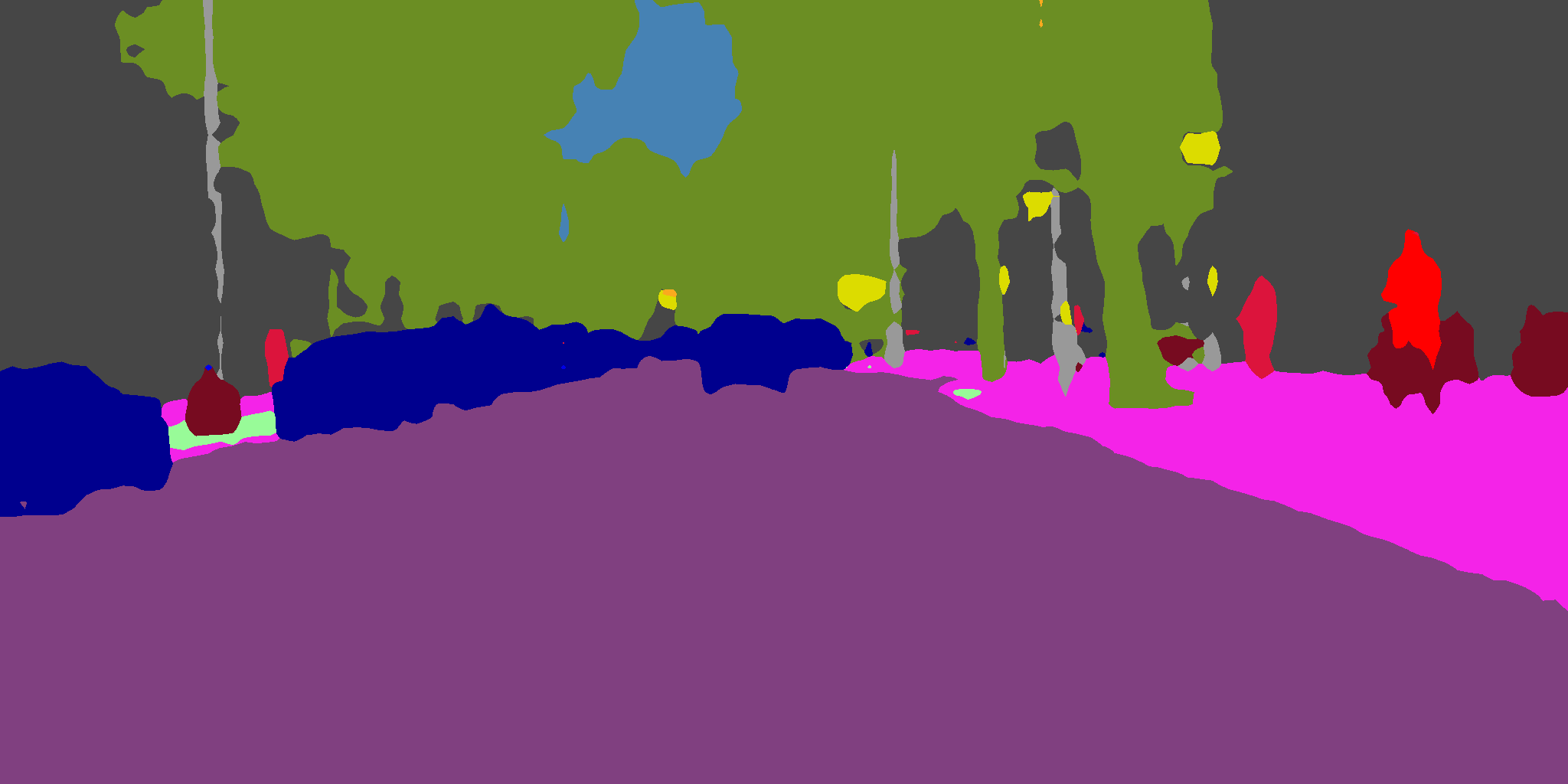}
\end{minipage}
\caption{
Domain adaptive segmentation for the task GTA5-to-Cityscapes: Columns (a) and (b) show a few sample images from the dataset Cityscapes and the ground truth of their pixel-level segmentation, respectively. Columns (c), (d) and (e) present the qualitative result that is produced by state-of-the-art methods CrCDA~\cite{huang2020contextual}, SVMin~\cite{guan2020scale} and the proposed MLAN, respectively. Best viewed in color and zoom in for details.
}
\label{fig:results_supple}
\end{figure*}

As shown in Figure \ref{fig:results_supple}, we present qualitative segmentation results on the adaptation task GTA5-to-Cityscapes. It is clear that the proposed method performs the best, especially in the local regions with class-transition ($i.e.$, regions that involve local context-relations among two or more classes). For example, as shown in the last row of Figure \ref{fig:results_supple}, CrCDA and SVMin falsely predict the trunk of the 'tree' as 'pole' while the proposed MLAN segments the same area correctly. We reckon the improvement is largely thanks to the proposed multi-level consistencies regularization that well rectifies region-level context-relation alignment ($e.g.$, the 'pole' is not supposed to be on the top of 'vegetation') as well as image-level feature alignment. 

\begin{figure}[t]
\centering
\includegraphics[width=.9\linewidth]{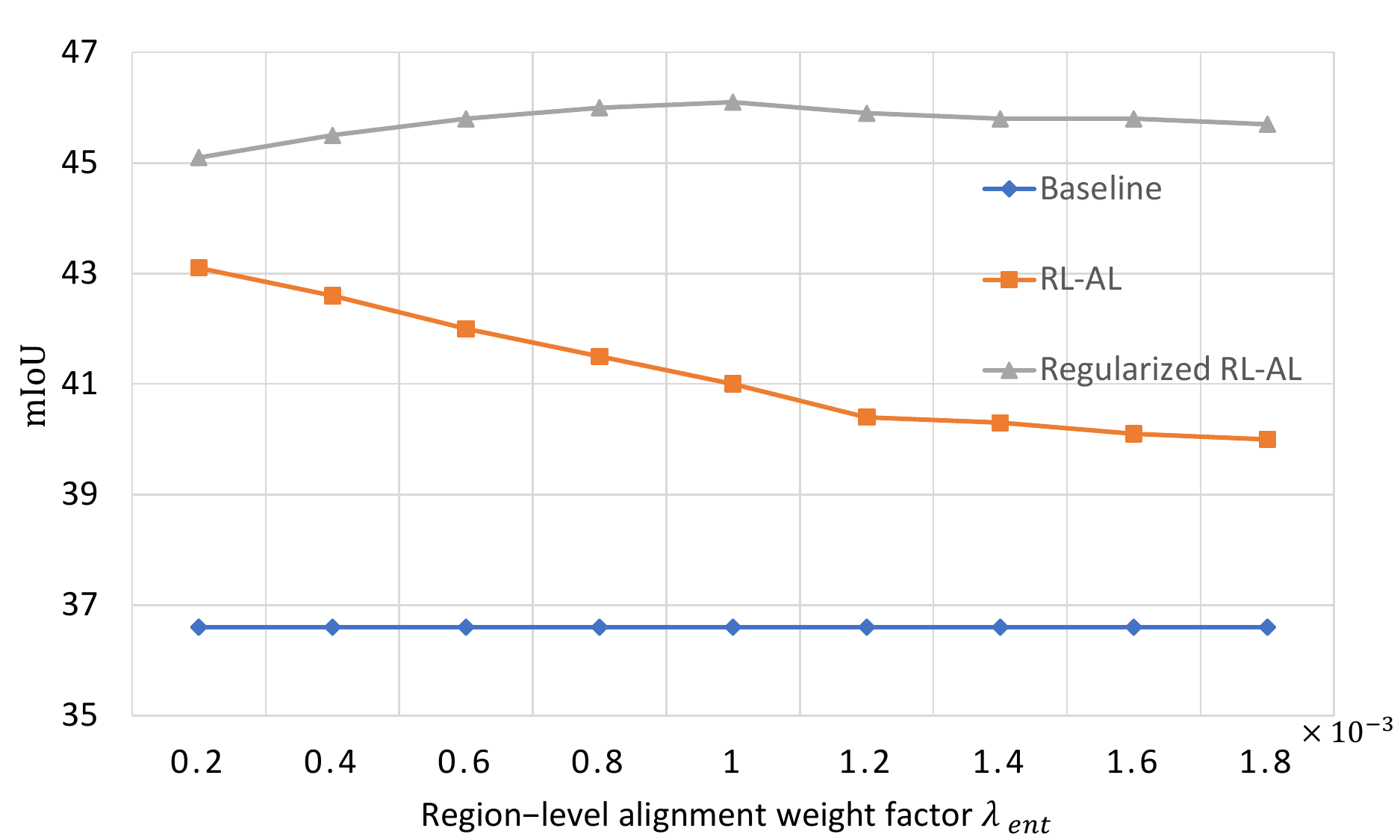}
\caption{
Parameter studies of region-level adversarial learning (RL-AL) weight factor $\lambda_{ent}$ (with ResNet-101): For the domain adaptive segmentation task GTA5-to-Cityscapes, the segmentation with the original RL-AL is sensitive to different weight factors, showing that its adversarial loss is an unstable objective. As a comparison, the proposed regularized RL-AL is much more stable and robust with different weight factors as the proposed regularization incorporates global image-level information.
}
\label{fig:weight1}
\end{figure}

\begin{figure}[t]
\centering
\includegraphics[width=.9\linewidth]{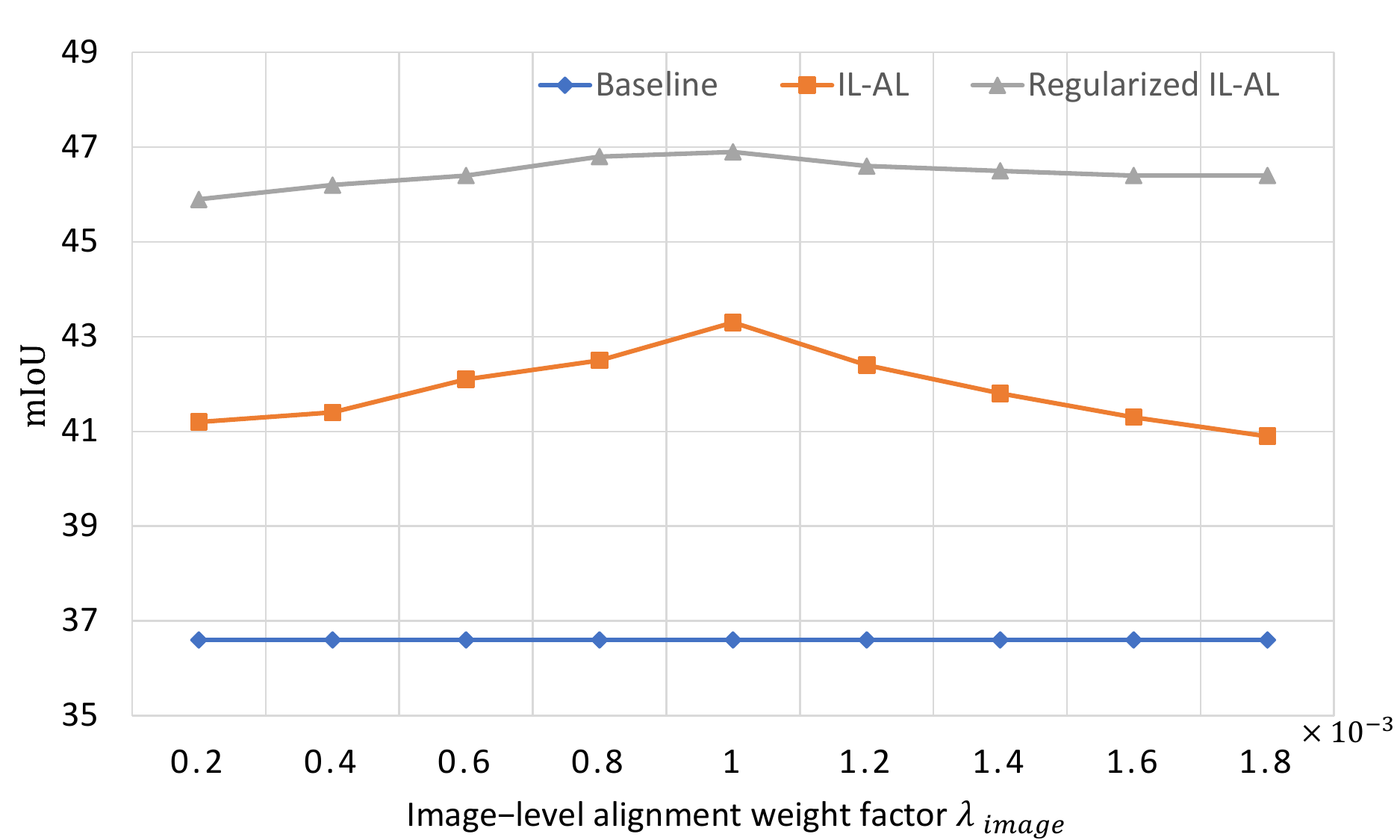}
\caption{
Parameter studies of global image-level adversarial learning (IL-AL) weight factor $\lambda_{image}$ (with ResNet-101): For the domain adaptive semantic segmentation task GTA5-to-Cityscapes, the segmentation of the original IL-AL is sensitive to different weight factors, showing that its adversarial loss is an unstable objective. As a comparison, the proposed regularized IL-AL is more stable and robust to different weight factors as the proposed regularization incorporates local region-level information.
}
\label{fig:weight2}
\end{figure}

\subsection{Parameter studies}
We analyze the influence of the weight factors that are employed to balance the adaptation/alignment objectives ($i.e.$, $\lambda_{ent}$ for region-level adversarial learning (RL-AL) and $\lambda_{image}$ for image-level adversarial learning) and other factors such as the source supervised segmentation loss and region-level learning loss. As shown in Figures \ref{fig:weight1} and \ref{fig:weight2}, the segmentation performance of proposed regularized RL-AL/IL-AL is insensitive and robust to a variety of weight factor values. Specifically, for RL-AL, the segmentation performance varies from $40.0\%$ to $43.1\%$ while its weight factor ranges from $0.2 \times 10^{-3}$ to $1.8 \times 10^{-3}$. On the contrary, the segmentation performance of proposed regularized RL-AL varies from $45.1\%$ to $46.1\%$ while its weight factor changes from $0.2 \times 10^{-3}$ to $1.8 \times 10^{-3}$. The IL-AL and the rgularized IL-AL show similar phenomenon during same weight factor variation, where IL-AL varies between $40.9\%$ to $43.3\%$ and IL-AL varies between $45.9\%$ to $46.9\%$.
The phenomenon described above indicates that the original adversarial loss ($i.e.$, RL-AL and IL-AL) is an unstable training objective, where the proposed regularization can effective improve it and make it more robust and stable.

\section{Conclusion}
In this work, we propose an innovative multi-level adversarial network (MLAN) for domain adaptive semantic segmentation, which aims to achieve both region-level and image-level cross-domain consistencies, optimally. MLAN introduces two novel designs, namely region-level adversarial learning (RL-AL) and co-regularized adversarial learning (CR-AL).
Specifically, RL-AL learns the prototypical region-level context-relations explicitly in the feature space of a labelled source domain and transfer them to an unlabelled target domain via adversarial learning. CR-AL integrates local region-level AL with global image-level AL and conducts mutual regularization. 
Furthermore, MLAN calculates a multi-level consistency map (MLCM) to guide the domain adaptation in input and output space. We experimentally evaluate the superior of the proposed method on the adaptation from a source domain of synthesized images to a target domain of real images. The proposed methods surpass state-of-the-arts by a large margin. In addition, we perform extensive ablation studies that illustrate the contribution of each design to the overall performance of the proposed method. We also conduct parameter studies and find that the general adversarial learning is an unstable objective for semantic segmentation and the proposed method can well rectify it. In the future work, we will explore more segmentation properties to make adaptation process/loss more relevant to the main task ($i.e.$, segmentation).

\section*{Acknowledgment}
This research was conducted at Singtel Cognitive and Artificial Intelligence Lab for Enterprises (SCALE@NTU), which is a collaboration between Singapore Telecommunications Limited (Singtel) and Nanyang Technological University (NTU) that is funded by the Singapore Government through the Industry Alignment Fund ‐ Industry Collaboration Projects Grant.


\bibliographystyle{plain}
\small\bibliography{egbib}

\end{document}